\documentclass[pmlr]{jmlr}

\usepackage{longtable}
\usepackage{booktabs}
\usepackage[load-configurations=version-1]{siunitx}
\usepackage{pgfplots}
\usepackage{pgfplotstable}
\usepgfplotslibrary{groupplots}
\usetikzlibrary{positioning, fit, arrows.meta, calc, plotmarks}
\usepackage[dvipsnames]{xcolor}
\definecolor{GlobalResidualConformalColor}{RGB}{31,119,180}
\pgfplotsset{compat=1.18}
\usepackage{pgfplots}
\usepackage{pgfplotstable}
\usepgfplotslibrary{fillbetween}

\usepackage{booktabs}
\usepackage{array}
\usepackage{amssymb}
\usepackage{multirow}
\usepackage{float}
\usepackage{placeins}
\newcommand{\cmark}{\checkmark}
\newcommand{\xmark}{\(\times\)}

\definecolor{cGlobalResidualConformal}{RGB}{31,119,180}
\definecolor{cHorizonWiseConformal}{RGB}{255,127,14}
\definecolor{cMaxScoreSplitConformal}{RGB}{44,160,44}
\definecolor{cResidualQuantileConformal}{RGB}{214,39,40}
\definecolor{cRollingHorizonConformal}{RGB}{148,103,189}
\definecolor{cRollingTrajectoryConformal}{RGB}{140,86,75}
\definecolor{cExponentiallyWeightedHorizonConformal}{RGB}{227,119,194}
\definecolor{cExponentiallyWeightedTrajectoryConformal}{RGB}{23,190,207}
\definecolor{cEnbpiHorizon}{RGB}{128,128,0}
\definecolor{cEnbpiTrajectory}{RGB}{85,107,47}
\definecolor{cAciHorizon}{RGB}{255,0,255}
\definecolor{cAciTrajectory}{RGB}{238,130,238}
\definecolor{cAgaciHorizon}{RGB}{255,215,0}
\definecolor{cAgaciTrajectory}{RGB}{184,134,11}
\definecolor{cBonferroniHorizonConformal}{RGB}{128,128,128}
\definecolor{cSidakHorizonConformal}{RGB}{192,192,192}
\definecolor{cCopulacpts}{RGB}{0,128,128}
\definecolor{cTraceConformal}{RGB}{0,0,0}
\definecolor{cGlobalCrc}{RGB}{0,0,128}
\definecolor{cHorizonProfileCrc}{RGB}{255,140,0}
\definecolor{cTrajectoryStratifiedCrc}{RGB}{0,100,0}
\definecolor{cTraceCrc}{RGB}{139,0,0}
\definecolor{cHybridHorizonAdaptiveTraceCrc}{RGB}{127,127,127}

\usepackage{tikz}
\usepgfplotslibrary{groupplots}
\usetikzlibrary{calc}

\definecolor{cGlobalResidualConformal}{RGB}{31,119,180}
\definecolor{cHorizonWiseConformal}{RGB}{255,127,14}
\definecolor{cMaxScoreSplitConformal}{RGB}{44,160,44}
\definecolor{cResidualQuantileConformal}{RGB}{214,39,40}
\definecolor{cRollingHorizonConformal}{RGB}{148,103,189}
\definecolor{cRollingTrajectoryConformal}{RGB}{140,86,75}
\definecolor{cExponentiallyWeightedHorizonConformal}{RGB}{227,119,194}
\definecolor{cExponentiallyWeightedTrajectoryConformal}{RGB}{23,190,207}
\definecolor{cEnbpiHorizon}{RGB}{128,128,0}
\definecolor{cEnbpiTrajectory}{RGB}{85,107,47}
\definecolor{cAciHorizon}{RGB}{255,0,255}
\definecolor{cAciTrajectory}{RGB}{238,130,238}
\definecolor{cAgaciHorizon}{RGB}{255,215,0}
\definecolor{cAgaciTrajectory}{RGB}{184,134,11}
\definecolor{cBonferroniHorizonConformal}{RGB}{128,128,128}
\definecolor{cSidakHorizonConformal}{RGB}{192,192,192}
\definecolor{cCopulacpts}{RGB}{0,128,128}
\definecolor{cTraceConformal}{RGB}{0,0,0}
\definecolor{cGlobalCrc}{RGB}{0,0,128}
\definecolor{cHorizonProfileCrc}{RGB}{255,140,0}
\definecolor{cTrajectoryStratifiedCrc}{RGB}{0,100,0}
\definecolor{cTraceCrc}{RGB}{139,0,0}
\definecolor{cHybridHorizonAdaptiveTraceCrc}{RGB}{127,127,127}

\definecolor{pairEWColor}{RGB}{31,119,180}
\definecolor{pairEnbPIColor}{RGB}{255,127,14}
\definecolor{pairACIColor}{RGB}{44,160,44}
\definecolor{pairAgACIColor}{RGB}{214,39,40}
\definecolor{cTraceCrcColor}{RGB}{148,103,189}

\tikzset{
    pairEW/.style={color=pairEWColor},
    pairEnbPI/.style={color=pairEnbPIColor},
    pairACI/.style={color=pairACIColor},
    pairAgACI/.style={color=pairAgACIColor},
    cTraceCrc/.style={color=cTraceCrcColor}
}

\pgfplotsset{
  conformalAxis/.style={
    width=0.92\linewidth,
    height=0.50\linewidth,
    grid=both,
    major grid style={draw=black!12},
    minor grid style={draw=black!6},
    tick align=outside,
    tick style={black!60},
    label style={font=\small},
    tick label style={font=\small},
    legend style={font=\scriptsize, draw=none, fill=none, /tikz/every even column/.append style={column sep=0.4em}},
  }
}

\pgfplotsset{
  discard if not/.style 2 args={
    x filter/.append code={
      \edef\tempa{\thisrow{#1}}
      \edef\tempb{#2}
      \ifx\tempa\tempb
      \else
        
      \fi
    }
  }
}
\jmlrvolume{329}
\jmlryear{2026}
\jmlrworkshop{Conformal and Probabilistic Prediction with Applications}
\jmlrproceedings{PMLR}{Proceedings of Machine Learning Research}

\title[TRACE-CRC for Multi-Step CSI Prediction]{TRACE-CRC: Trajectory-Adaptive Conformal Risk Control for Multi-Step Channel State Information Prediction}
\author{%
\Name{Kiarash Rezaei} \Email{kiarashr@chalmers.se}\\
\Name{Mehdi Sattari} \Email{mehdi.sattari@chalmers.se}\\
\Name{Javad Aliakbari} \Email{javada@chalmers.se}\\
\Name{Tommy Svensson} \Email{tommy.svensson@chalmers.se}\\
\Name{Paolo Monti} \Email{mpaolo@chalmers.se}\\
\Name{Carlos Natalino} \Email{carlos.natalino@chalmers.se}\\
\addr Department of Electrical Engineering, Chalmers University of Technology, Gothenburg, Sweden
}

\editor{Ernst Ahlberg, Ulf Johansson, Henrik Boström, Alberto Carlevaro, Johan Hallberg Szabadváry and Lars Carlsson}

\begin{document}

\maketitle

\begin{abstract}
Reliable prediction of time-varying channel state information (CSI) is essential for efficient wireless communication. Each CSI frame is a matrix-valued representation of the wireless channel response, and a sequence of CSI frames forms a temporal channel trajectory. Modern deep learning-based CSI predictors, however, often provide only point predictions and lack calibrated uncertainty estimates. This limitation is particularly problematic in multi-step CSI prediction, where the target is a sequence of future CSI matrices, and downstream decisions such as beamforming or scheduling may fail if any part of the predicted trajectory is unreliable. We propose trajectory-adaptive calibration and error profiling with conformal risk control (TRACE-CRC), a method for trajectory-aware uncertainty quantification in multi-step CSI prediction. TRACE-CRC constructs Frobenius-norm uncertainty balls around predicted CSI matrices and controls the risk that at least one future frame is uncovered. Instead of calibrating each future step independently, TRACE-CRC combines future-step-dependent error profiling, trajectory difficulty stratification, and learn-then-test (LTT) risk control. Empirically, TRACE-CRC achieves reliable trajectory-level coverage with substantially smaller uncertainty balls than conservative multi-step corrections, while avoiding the trajectory undercoverage of compact stepwise and adaptive conformal baselines.
\end{abstract}


\section{Introduction}
\label{sec:introduction}
Recent advances in wireless technologies, such as massive multiple-input multiple-output (MIMO), have significantly improved the efficiency of wireless communication systems.
In a MIMO system, multiple antennas are used at the transmitter and receiver so that several signal paths can be exploited simultaneously to improve data rate and reliability. Fully realizing these gains requires accurate channel state information (CSI), a high-dimensional complex-valued representation of the wireless channel response. CSI characterizes the strength and phase of the propagation paths and supports transmission decisions such as beamforming, precoding, and resource allocation.
The ability to fully exploit these technologies critically depends on the accurate acquisition of CSI, a high-dimensional complex-valued representation of the wireless channel response. 
However, acquiring CSI remains challenging due to the high dimensionality, complex propagation environment, and multi-modality of wireless channels \citep{massivemimobook}.

The conventional approach for CSI acquisition relies on pilot transmission and channel estimation at the receiver. CSI prediction provides an alternative: future channel states are forecast from past observations, helping mitigate channel aging and reduce pilot overhead, which are key bottlenecks in next-generation wireless systems \citep{10422880,9210016}. Traditional CSI prediction methods include Kalman filtering \citep{9210016}, autoregressive models \citep{1512123}, and linear extrapolation \citep{9127447}. While analytically tractable, these methods often fail to capture the complex nonlinear dynamics of wireless channels.

Recent deep learning methods, including recurrent neural networks \citep{8746352}, transformer-based architectures \citep{10965849}, and diffusion models \citep{sattari2025csipredictionusingdiffusion}, have significantly improved CSI prediction accuracy. Nevertheless, deep learning-based CSI predictors typically produce point estimates without calibrated uncertainty information. This limits reliability in deployment, since downstream decisions made by the wireless control plane, such as beamforming, scheduling, and link adaptation, depend not only on the predicted channel but also on the confidence associated with that prediction. In the multi-step setting considered here, the predictor outputs several future CSI frames; we refer to each future prediction step as a forecast horizon, or simply a horizon, and to the full sequence as a predicted CSI trajectory.

Conformal prediction (CP) provides a model-agnostic framework for constructing calibrated prediction sets around black-box model outputs \citep{vovk/1062391,angelopoulos2022gentleintroductionconformalprediction}. This makes CP an appropriate candidate for uncertainty quantification in CSI prediction: instead of replacing the predictor, CP can wrap an existing CSI prediction model with uncertainty regions. Yet, multi-step CSI prediction poses challenges that standard split CP does not fully address. In other words, the prediction target is a trajectory of high-dimensional complex CSI matrices; errors can grow across the forecast horizon, and different channel-evolution patterns can exhibit varying levels of predictability. Moreover, for a wireless deployment, reliability should be assessed at the trajectory level, where a predicted CSI trajectory may be unreliable if any future CSI frame falls outside its uncertainty region.

Existing CP methods for wireless communications have mainly focused on classification, robust optimization, context-shift calibration, or uncertainty for specific channel prediction pipelines \citep{10262367}. In parallel, the broader time-series CP literature has developed tools for sequential adaptation, horizon dependence, and multi-step prediction, while conformal risk control (CRC) provides mechanisms for calibrating predictive systems under user-specified losses \citep{Angelopoulos:22CRC}. However, existing methods do not directly target the combination of matrix-valued CSI prediction, multi-step trajectory-level reliability, and explicit risk control certification considered in this work.

This paper proposes \emph{TRACE-CRC}, a trajectory-adaptive CRC framework for multi-step CSI prediction. It constructs Frobenius-norm uncertainty balls around predicted CSI matrices, adapts their radii using horizon-dependent error profiles and trajectory-level difficulty strata, and certifies trajectory-level risk using Learn-then-Test (LTT). 
The resulting framework aims to provide calibrated uncertainty estimates for the complete predicted CSI trajectory rather than only for isolated forecast horizons,  shifting the calibration target from marginal per-step coverage to control of the full-trajectory failure event. This trajectory-level perspective supports more informed downstream wireless decisions, where a single unreliable future CSI frame can affect beamforming, scheduling, precoding, or link adaptation across the predicted trajectory. The source code and scripts required to reproduce the experiments are publicly available at
\url{https://github.com/kiarashRezaei/trace-crc}.

The main contributions of this work are summarized as follows:
\begin{itemize}
    \item We formulate multi-step CSI reliability as a trajectory-level CRC problem, where a predicted CSI trajectory is considered unreliable if at least one future CSI matrix falls outside its uncertainty ball.

    \item We extend structured multi-step conformal prediction to matrix-valued CSI trajectories by constructing per-horizon Frobenius-norm uncertainty balls whose radii adapt to both horizon-dependent error growth and heterogeneous trajectory difficulty.

    \item We combine trajectory-adaptive conformal calibration with an LTT risk control layer that selects an uncertainty rule with a finite-sample trajectory-level risk control.

    \item We evaluate TRACE-CRC against horizon-wise, trajectory-level, online adaptive, weighted, joint multi-step, and conformal risk control baselines. TRACE-CRC achieves trajectory coverage \(0.933\) at target coverage \(0.90\), while substantially reducing the conservatism of simultaneous-correction baselines.
\end{itemize}

\section{Related Work}
\label{sec:related_work}

\subsection{Conformal Prediction for Marginal Coverage}
\label{subsec:related_marginal_cp}
Classical split conformal methods provide finite-sample marginal coverage under the exchangeability assumption by calibrating a nonconformity score on held-out data \citep{papadopoulos2002inductive,lei2018distribution}. Several extensions improve adaptivity or robustness under more complex data settings. Conformalized quantile regression (CQR) combines quantile regression with conformal calibration to obtain adaptive regression intervals \citep{romano2019conformalized}. Weighted conformal methods address covariate shift and more general forms of non-exchangeability by reweighting calibration residuals \citep{tibshirani2019conformal,barber2023conformal}. These methods form the foundation for uncertainty calibration, but they primarily target marginal prediction-set validity rather than trajectory-level reliability over multiple forecast horizons.

\subsection{Conformal Prediction for Time Series and Multi-Step Forecasting}
\label{subsec:related_timeseries_cp}
Sequential and time-series prediction often violate the exchangeability assumptions underlying classical CP. This has motivated the development of conformal methods for temporally dependent, nonstationary, and distribution-shifting data. Ensemble batch prediction intervals (EnbPI) constructs bootstrap-based prediction intervals for dynamic time series and updates residuals sequentially during deployment \citep{xu2021conformal,xu2023conformal}. Adaptive conformal inference (ACI) updates the miscoverage level online to maintain long-run coverage under distribution shift \citep{gibbs2021adaptive}, with related adaptive variants, aggregated adaptive conformal inference (AgACI), developed for time-series forecasting \citep{zaffran2022adaptive}. Other approaches exploit temporal structure in residuals or nonconformity scores: sequential predictive conformal inference (SPCI) estimates future residual quantiles using sequential residual quantile regression \citep{xu2023sequential}, while kernel-based optimally weighted conformal prediction intervals (KOWCPI) uses kernel-based optimally weighted residual quantile regression for non-exchangeable time series \citep{lee2025kernel}. Multi-step forecasting introduces an additional challenge because uncertainty must be calibrated across multiple forecast horizons. CopulaCPTS models dependence among horizon-wise conformity scores through an empirical copula to obtain valid multi-step uncertainty regions while avoiding overly conservative Bonferroni corrections \citep{sun2024copula}; related multi-step conformal approaches include ConForME and online multi-step conformal forecasting methods \citep{galvaolopes2024conforme,wang2024online}.Several of these methods are adapted as baselines in our experiments where adaptive time-series methods are evaluated through separate horizon-wise and trajectory-wise variants, while multi-step methods address dependence across horizons without CRC-style risk certification. TRACE-CRC combines horizon-dependent structure and trajectory-level CRC certification in one framework.

\subsection{Conformal Risk Control}
\label{subsec:related_crc}
Beyond coverage guarantees, conformal risk control (CRC) calibrates prediction sets with respect to user-specified losses. Distribution-free risk-controlling prediction sets provide finite-sample control of expected losses for set-valued predictions \citep{bates2021distributionfree}. The LTT framework casts calibration as a multiple-testing problem and provides high-confidence risk control for general predictive algorithms \citep{angelopoulos2022learntestcalibratingpredictive}. CRC further specializes this perspective to monotone losses and generalizes split CP from coverage control to broader risk control \citep{Angelopoulos:22CRC}. Recent work has extended risk control ideas to non-exchangeable data \citep{farinhas2024nonexchangeable}, localized and online guarantees \citep{zecchin2024localized}, fairness-aware calibration \citep{zhang2024fair}, and application-oriented risk control for model alignment and efficient inference \citep{overman2024aligning,jazbec2024fast}. A closely related line of work applies CRC to sequential prediction and control, including wireless networking applications \citep{zecchin2024forking}. TRACE-CRC instead applies CRC to matrix-valued multi-step CSI prediction with a trajectory-level failure loss.

\subsection{Conformal Prediction in Wireless Communications}
\label{subsec:related_wireless_cp}
CP has recently been explored in several wireless communication settings. Standard split CP and cross-validation-based CP have been used for wireless classification tasks such as few-shot demodulation and modulation classification \citep{10096780,10262367}. Online CP has been studied for channel prediction in wireless systems \citep{10262367}. To address the training-deployment mismatch, \citep{11329107} proposes a meta-learned weighted CP method for context-dependent covariate shift. For decision-oriented wireless optimization, \citep{11143438} use split CP to construct channel uncertainty sets for robust beamforming. Most closely related to CSI prediction, \citep{kim2026mimochannelpredictiondeep} use CQR within a deep conformal Bayes filter to obtain calibrated uncertainty for MIMO channel prediction. A broader overview of CP-based calibration for wireless AI is provided by \citet{simeone2025conformalcalibrationensuringreliability}.
Compared with these works, TRACE-CRC differs in both the prediction object and the reliability target. It calibrates Frobenius-norm uncertainty balls for matrix-valued multi-step CSI trajectories and certifies trajectory-level failure risk, rather than focusing on classification, context-shift calibration, robust decision sets, or uncertainty for individual channel predictions.
\section{Dataset and Problem Formulation}
\label{sec:preliminaries}
We study conformal uncertainty quantification for multi-step CSI prediction. In this setting, a pretrained CSI predictor observes past CSI frames and outputs a future CSI trajectory over multiple forecast horizons. Since downstream wireless decisions may depend on the consistency of the predicted channel evolution, our goal is not only to assess point-prediction accuracy, but also to construct Frobenius-norm uncertainty balls around predicted CSI matrices that provide trajectory-level reliability. We first describe the CSI dataset and prediction model, and then formalize the corresponding trajectory-level conformal prediction objective.

\subsection{Dataset}
\label{sec:dataset}

We consider CSI trajectories generated from an underlying channel distribution. 
Informally, this distribution describes how the wireless channel evolves over time under
random propagation conditions such as mobility, scattering, and multipath propagation effects. A CSI
frame at time \(t\) is a complex-valued matrix
\[
    \mathbf{H}_t \in \mathbb{C}^{N_{\mathrm t}\times N_{\mathrm c}},
\]
where \(N_{\mathrm t}\) is the antenna dimension and \(N_{\mathrm c}\) is the subcarrier or
frequency dimension. Thus, each CSI frame can be viewed as a high-dimensional matrix-valued
observation of the channel at one time instant.

The finite dataset consists of sampled realizations of this channel process
\[
    \mathcal{D}
    =
    \left\{
    \left(
    \mathbf{H}^{(i)}_{1},
    \mathbf{H}^{(i)}_{2},
    \ldots,
    \mathbf{H}^{(i)}_{T_i}
    \right)
    \right\}_{i=1}^{\mathrm{M}},
\]
where \(i\) indexes trajectories, \(T_i\) is the length of trajectory \(i\), \(t\)
indexes time within a trajectory, and \(\mathrm{M}\) is the number of trajectories.
We use the CSI dataset generation setup from \citet{sattari2025csipredictionusingdiffusion}. The dataset consists of 1{,}000 independent CSI samples, each with $N_t=16$ transmit antennas and $N_c=16$ subcarriers, operating at a carrier frequency of 28~GHz. User velocities are uniformly sampled between 30 and 120~km/h, while the channel model is randomly selected from the 3GPP-compliant CDL models (\texttt{CDL-A}--\texttt{CDL-E}) \citep{3gpp.38.901}. Each sample contains \(100\) time steps (frames), with consecutive CSI frames separated by approximately \(33.3\,\mu\mathrm{s}\). Since the CSI is complex-valued, the real and imaginary components are stored as separate channels. Consequently, each sample is represented as a tensor of shape $100 \times 2 \times 16 \times 16$, where the first dimension corresponds to the 100 time steps, the second dimension represents the real and imaginary components of the CSI, and the remaining two dimensions correspond to the antenna and subcarrier dimensions, respectively.

\begin{figure}[t]
    \centering
    \includegraphics[width=0.20\textwidth]{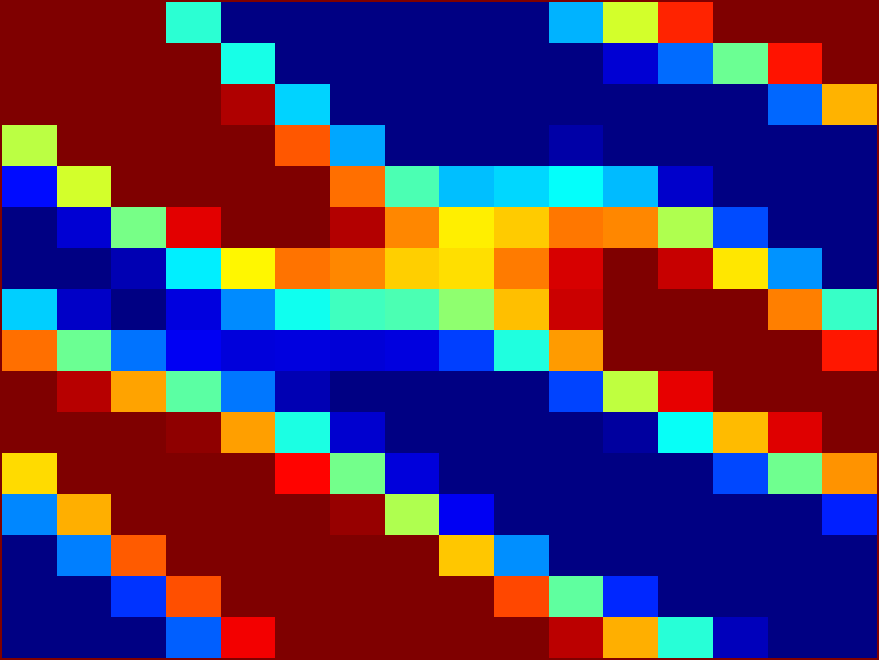}
    \hspace{0.01\textwidth}
    \includegraphics[width=0.20\textwidth]{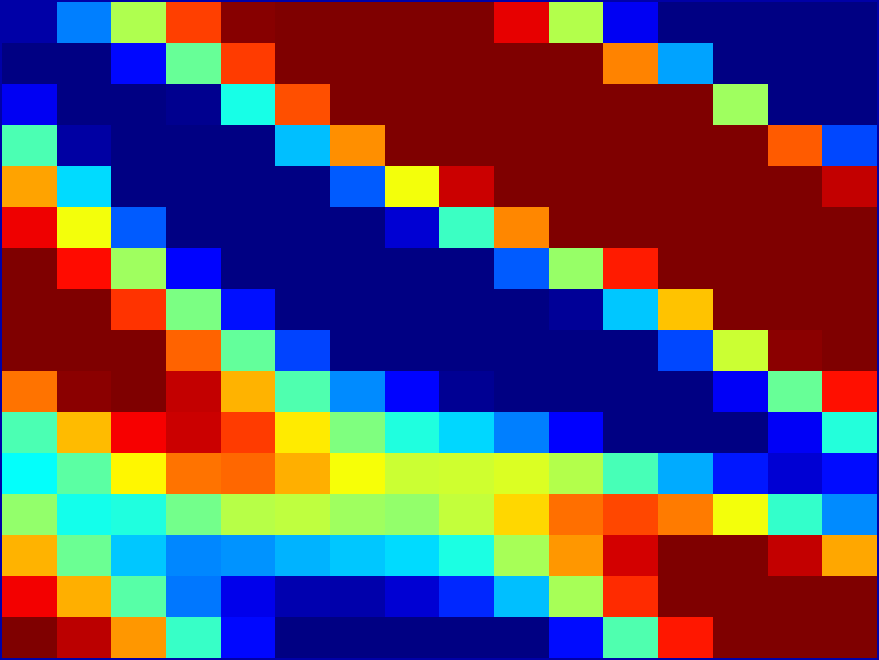}
    \hspace{0.02\textwidth}
    \raisebox{2.5\height}{\Huge $\cdots$}
    \hspace{0.02\textwidth}
    \includegraphics[width=0.20\textwidth]{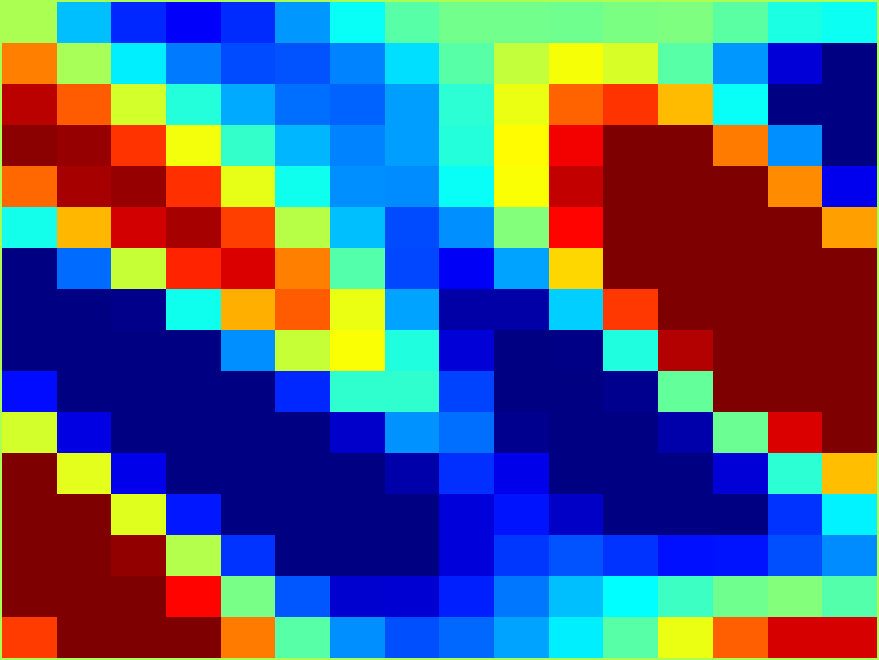}
    \caption{Snapshots of the real part of a CSI frame at different time steps.}
    \label{fig:csi_snapshots}
\end{figure}

The exchangeability assumption applies across complete CSI trajectories, not across frames within a trajectory. Hence, temporal dependence and channel evolution within each trajectory are allowed. In our simulation, trajectories are generated independently using the same channel-generation procedure and parameter distributions, making trajectory-level exchangeability reasonable in this setting.
Figure~\ref{fig:csi_snapshots} shows example snapshots from one trajectory by visualizing the real part of the CSI matrices at different time horizons. The variation across snapshots illustrates the temporal evolution of the wireless channel.

\subsection{CSI Prediction}
\label{subsec:csi_prediction}
Time-varying CSI prediction can be viewed as a matrix-valued time-series forecasting problem.
Let \(\mathbf{H}_{\mathrm p}\) denote the past CSI frames used as input, and let
\(\mathbf{H}_{\mathrm f}\) denote the future CSI frames to be predicted
\[
\mathbf{H}_{\mathrm{p}}
=
\left(
    \mathbf{H}_{t-N_{\mathrm p}+1},
    \mathbf{H}_{t-N_{\mathrm p}+2},
    \ldots,
    \mathbf{H}_{t}
\right),
\qquad
\mathbf{H}_{\mathrm{f}}
=
\left(
    \mathbf{H}_{t+1},
    \mathbf{H}_{t+2},
    \ldots,
    \mathbf{H}_{t+N_{\mathrm f}}
\right).
\]
Here, \(\mathbf H_{\mathrm p}\) and \(\mathbf H_{\mathrm f}\) are random variables induced by
the underlying channel distribution, while the dataset in Subsection~\ref{sec:dataset}
contains sampled realizations of these quantities.

Let \(\mathcal G\) denote the class of admissible predictors mapping past CSI trajectories to future CSI trajectories. Under the mean squared error criterion, the population-optimal predictor $g^\ast(\mathbf{H}_{\mathrm{p}})$ is
\begin{equation}
    g^\ast(\mathbf{H}_{\mathrm{p}})
    =
    \arg\min_{g\in \mathcal G}
    \mathbb{E}
    \left[
    \left\lVert
    \mathbf{H}_{\mathrm{f}} - g(\mathbf{H}_{\mathrm{p}})
    \right\rVert^2
    \right].
\end{equation}
The optimal prediction rule is the conditional mean estimator
\begin{equation}
\label{eq:CME}
    g^\ast(\mathbf{H}_{\mathrm{p}})
    =
    \mathbb{E}
    \left[
    \mathbf{H}_{\mathrm{f}}
    \mid
    \mathbf{H}_{\mathrm{p}}
    \right] 
    =
    \int
    \mathbf{h}_{\mathrm f}
    \,
    \mathbb{P}
    \left(
    \mathbf{h}_{\mathrm f}
    \mid
    \mathbf{H}_{\mathrm p}
    \right)
    d\mathbf{h}_{\mathrm f}.
\end{equation}

In practice, computing the minimum mean square error (MMSE) predictor is challenging because the conditional law of \(\mathbf H_{\mathrm f}\) given \(\mathbf H_{\mathrm p}\), denoted by \(\mathbb{P}(\mathbf H_{\mathrm f}\mid \mathbf H_{\mathrm p})\), is unknown and high-dimensional. A deep learning-based predictor \(g_{\theta}\) trained with mean square error (MSE) can be viewed as a data-driven approximation to this conditional mean. However, deterministic predictors do not explicitly capture the stochastic and potentially multi-modal evolution of future CSI trajectories.

Recent diffusion-based CSI predictors address this limitation by modeling future CSI generation probabilistically. Following \citet{sattari2025csipredictionusingdiffusion}, we use a pretrained predictor based on a temporal encoder and diffusion generator. The temporal encoder extracts representations from the past CSI frames, while the diffusion
generator produces future CSI samples. At inference time, the model is used autoregressively: each predicted CSI frame is fed back as input for predicting the next frame.

For a sampled trajectory \(i\), the past CSI input is
\(
    \mathbf H^{(i)}_{\mathrm p}
    =
    \left(
    \mathbf H^{(i)}_{t-N_{\mathrm p}+1},
    \ldots,
    \mathbf H^{(i)}_{t}
    \right).
\)
Let \(g_\theta\) denote the trained diffusion-based CSI predictor, where \(\theta\) represents the learned parameters of the temporal encoder and diffusion generator. At inference time, \(g_\theta\) generates a predicted future CSI trajectory from the past CSI window
\[
    \widehat{\mathbf H}^{(i)}_{\mathrm f}
    =
    g_\theta
    \left(
    \mathbf H^{(i)}_{\mathrm p}
    \right)
    =
    \left(
    \widehat{\mathbf H}^{(i)}_{t+1},
    \ldots,
    \widehat{\mathbf H}^{(i)}_{t+N_{\mathrm f}}
    \right).
\]

While the autoregressive scheme provides a flexible prediction framework, it also induces horizon-dependent uncertainty. Prediction errors can propagate through the feedback loop and accumulate over future steps, so the reliability of the predicted CSI trajectory may deteriorate as the forecast horizon increases. This motivates the conformal uncertainty quantification problem considered next.

\subsection{Conformal Problem Formulation}
\label{sec:problem_formulation}
Building on the multi-step CSI prediction setup in Subsection~\ref{subsec:csi_prediction}, we now formulate trajectory-level conformal uncertainty quantification. Each future index \(j\in\{1,\dots,N_{\mathrm f}\}\) is a forecast horizon, and the collection of all \(N_{\mathrm f}\) future CSI frames forms the predicted CSI trajectory.

For trajectory $i$ and forecast horizon $j$, we define the Frobenius residual as
\begin{equation}
\label{eq:residual_f_norm}
    E_{i,j}
    =
    \left\|
    \mathbf H^{(i)}_{t+j}
    -
    \widehat{\mathbf H}^{(i)}_{t+j}
    \right\|_F,
    \qquad
    j=1,\dots,N_{\mathrm f}.
\end{equation}

The Frobenius residual is well-suited for CSI uncertainty quantification because each CSI frame is a high-dimensional complex-valued matrix whose entries jointly describe the channel response across the spatial-frequency domain. 
Thus, a Frobenius-norm radius defines an uncertainty ball around the predicted CSI matrix in the full spatial-frequency channel space
\begin{equation}
\label{eq:f_ball}
    \mathcal{B}_{i,j}(r_{i,j})
    =
    \left\{
    \mathbf{G}\in\mathbb{C}^{N_{\mathrm t}\times N_{\mathrm c}}:
    \left\|
    \mathbf{G}
    -
    \widehat{\mathbf{H}}^{(i)}_{t+j}
    \right\|_F
    \le
    r_{i,j}
    \right\}.
\end{equation}
Here, \(\mathbf{G}\) is a generic candidate CSI matrix in the same spatial-frequency space as \(\mathbf{H}^{(i)}_{t+j}\). The ball is centered at the predicted CSI frame \(\widehat{\mathbf H}^{(i)}_{t+j}\). 
The radius \(r_{i,j}\) may depend on the forecast horizon and on quantities computed from the past CSI input and the predicted future CSI trajectory, but not on the unobserved true future CSI frame.

Let \(A_{i,j}\) denote the event that trajectory \(i\) is covered at forecast horizon \(j\)
\begin{equation}
    A_{i,j}
    =
    \left\{
    \mathbf{H}^{(i)}_{t+j}
    \in
    \mathcal{B}_{i,j}(r_{i,j})
    \right\}
    =
    \left\{
    E_{i,j}
    \le
    r_{i,j}
    \right\}.
\end{equation}
Thus, membership in the Frobenius ball is equivalent to the residual being no larger than the assigned radius.

Let $\alpha_{h}\in(0,1)$ denote the target horizon-wise miscoverage level. A horizon-wise objective marginally controls these events.
\begin{equation}
    \mathbb{P}(A_{i,j})
    \ge
    1-\alpha_{h},
    \qquad
    j=1,\dots,N_{\mathrm{f}}.
\end{equation}
However, marginal horizon-wise coverage does not generally imply that the entire predicted CSI trajectory is covered.
We therefore define the trajectory-level coverage event as
\(
    \mathcal C_i^{\mathrm{traj}}
    =
    \bigcap_{j=1}^{N_{\mathrm f}} A_{i,j}.
\)
Its complement corresponds to at least one uncovered horizon
\[
    \left(\mathcal C_i^{\mathrm{traj}}\right)^c
    =
    \left\{
    \exists j\in\{1,\dots,N_{\mathrm f}\}:
    E_{i,j}>r_{i,j}
    \right\}.
\]
The population trajectory-level risk is
\begin{equation}
\label{eq:traj_risk}
    \mathcal R_{\mathrm{traj}}
    =
    \mathbb P
    \left(
    \left(\mathcal C_i^{\mathrm{traj}}\right)^c
    \right)
    =
    \mathbb P
    \left(
    \exists j\in\{1,\dots,N_{\mathrm f}\}:
    E_{i,j}>r_{i,j}
    \right).
\end{equation}
Equivalently, the trajectory failure indicator is defined as
\begin{equation}
\label{eq:traj_failure_indicator}
    L_i^{\mathrm{traj}}
    =
    \mathbf 1
    \left\{
    \exists j\in\{1,\dots,N_{\mathrm f}\}:
    E_{i,j}>r_{i,j}
    \right\}.
\end{equation}
Since the population trajectory-level risk (\(\mathcal R_{\mathrm{traj}}\)) is unknown, we formulate the goal as a CRC problem. The calibration procedure returns a data-dependent uncertainty rule \(\widehat{\mathcal B}\), which assigns radii \(\widehat r_{i,j}\) using only the past CSI input and the predicted future CSI trajectory. Under trajectory-level exchangeability between the data used for risk control selection and future deployment trajectories, we seek a procedure such that
\begin{equation}
\label{eq:crc_goal}
    \mathbb{P}
    \left(
    \mathcal{R}_{\mathrm{traj}}(\widehat{\mathcal B})
    \le
    \alpha
    \right)
    \ge
    1-\delta .
\end{equation}
Here, \(\alpha\) is the target trajectory failure level, \(\delta\) is the allowed probability of certifying an uncertainty rule whose trajectory risk exceeds \(\alpha\), and the outer probability is over the data used by the CRC procedure.
The guarantee requires exchangeability across complete CSI trajectories, as discussed in Section~\ref{sec:dataset}.
\section{TRACE-CRC: Trajectory-Adaptive Conformal Risk Control}
\label{sec:method}

\subsection{Method Overview}
\label{sec:method_overview}

The trajectory-level objective in \eqref{eq:crc_goal} requires a rule that the corresponding uncertainty ball cover the full future CSI trajectory, rather than isolated forecast horizons. A single-radius trajectory-level conformal method can be inefficient because it applies the same radius across all horizons, while copula-based approaches explicitly model dependence across horizons, as in CopulaCPTS~\citep{sun2024copula}. TRACE-CRC adopts a different approach: it avoids explicit dependence modeling and instead adapts the radius to both horizon profile and predicted trajectory difficulty.

For trajectory \(i\) assigned to difficulty group \(g_i\in\{0,1\}\), TRACE-CRC sets the radius at forecast horizon \(j\) as
\begin{equation}
    r_{i,j}^{\star}
    =
    r_{g_i,j}^{\star}
    =
    \lambda^\star q_{g_i} w_j,
    \qquad
    j=1,\dots,N_{\mathrm f},
\end{equation}
where \(w_j\) is the horizon difficulty profile, \(\{q_0,q_1\}\) are the two group-wise conformal quantiles, \(q_{g_i}\) selects the quantile for the assigned group, and \(\lambda^\star\) is the global multiplier selected by LTT risk control.

The components \(w_j\), \(\{q_g\}_{g\in\{0,1\}}\), and \(\lambda^\star\) are estimated using disjoint calibration splits. TRACE-CRC then returns the Frobenius uncertainty ball \(\mathcal B_{i,j}^{\star}\) in~\eqref{eq:f_ball} with radius \(r_{i,j}^{\star}\). A future CSI trajectory is covered if every true future frame lies inside its corresponding horizon-wise uncertainty ball.

\subsection{Three-Way Calibration Split}
\label{subsec:three_way_split}
After splitting the available trajectory samples into calibration and test sets, TRACE-CRC further partitions the calibration CSI trajectories into three disjoint index subsets,
\[
\mathcal D_{\mathrm{cal}}
=
\mathcal D_{\mathrm{prof}}
\;\dot\cup\;
\mathcal D_{\mathrm{cp}}
\;\dot\cup\;
\mathcal D_{\mathrm{val}} .
\]
Thus, \(i\in\mathcal D_{\mathrm{prof}}\), \(i\in\mathcal D_{\mathrm{cp}}\), or \(i\in\mathcal D_{\mathrm{val}}\) indicates that trajectory \(i\) is assigned to the corresponding stage. For each calibration trajectory \(i\), we obtain the multi-step prediction from the predictor described in Subsection~\ref{subsec:csi_prediction} and compute the Frobenius residuals according to~\eqref{eq:residual_f_norm}. These residuals provide the calibration scores used by TRACE-CRC.

The profile subset \(\mathcal D_{\mathrm{prof}}\) is used to estimate the horizon difficulty profile \(w_j\) and to fit the trajectory difficulty model that assigns each trajectory \(i\) to a group \(g_i\in\{0,1\}\). The conformal calibration subset \(\mathcal D_{\mathrm{cp}}\) is used to compute the group-wise conformal quantiles \(q_g\). Finally, the validation subset \(\mathcal D_{\mathrm{val}}\) is reserved for LTT risk control selection of the global multiplier \(\lambda^\star\). Because the three subsets are disjoint, the horizon profile, grouping rule, and group-wise quantiles are fixed before the LTT hypothesis tests are applied on \(\mathcal D_{\mathrm{val}}\). This separation keeps the profiling, conformal calibration, and LTT testing stages statistically distinct, simplifying the validity argument. However, it may reduce sample efficiency when calibration trajectories are scarce. Cross-fitted or cross-conformal variants could improve data reuse, although retaining the same LTT risk-control guarantee would require additional analysis.

\subsection{Horizon Difficulty Profile}
\label{subsec:horizon_profile}

Prediction errors in multi-step CSI forecasting are horizon dependent, often increasing or changing shape across future horizons. TRACE-CRC therefore estimates a relative horizon difficulty profile \(w_1,\dots,w_{N_{\mathrm f}}\) using the profile split \(\mathcal D_{\mathrm{prof}}\).

For each forecast horizon \(j\), we first compute a raw horizon difficulty estimate as the empirical upper-quantile residual
\[
    w_j^{\mathrm{raw}}
    =
    Q_{1-\alpha_{h}}
    \left(
    \{E_{i,j}: i\in\mathcal D_{\mathrm{prof}}\}
    \right).
\]
where \(Q_{1-\alpha_h}\) denotes the empirical \((1-\alpha_h)\)-quantile. The parameter \(\alpha_h\) is used only to estimate the relative horizon difficulty profile and is distinct from the final trajectory-level risk target \(\alpha\).
Since horizon-wise empirical quantiles can fluctuate under finite calibration data, we smooth the raw profile by a local moving average. Specifically,
\[
    \widetilde w_j
    =
    \frac{1}{|\mathcal N_K(j)|}
    \sum_{k\in\mathcal N_K(j)}
    w_k^{\mathrm{raw}},
\]
where
\(
    \mathcal N_K(j)
    =
    \left\{
    k\in\{1,\dots,N_{\mathrm f}\}:
    |k-j|\le \left\lfloor K/2 \right\rfloor
    \right\}.
\)
In experiments, \(K=3\), so each horizon is smoothed using itself and its immediate neighbors, with boundary horizons averaged over the available terms.
We then stabilize the profile by imposing the lower floor
\(
c_{\min}
=
\rho\cdot\mathrm{median}(\widetilde w_1,\dots,\widetilde w_{N_{\mathrm f}}),
\)
where \(\rho>0\) is a stabilization parameter. The final profile is normalized to unit mean
\begin{equation}
\label{eq:horizon_diff}
    w_j
    =
    \frac{
    \max(\widetilde w_j,c_{\min})
    }{
    N_{\mathrm f}^{-1}\sum_{\ell=1}^{N_{\mathrm f}}
    \max(\widetilde w_\ell,c_{\min})
    } .
\end{equation}
The normalized profile satisfies \(N_{\mathrm f}^{-1}\sum_{j=1}^{N_{\mathrm f}}w_j=1\). Values \(w_j>1\) correspond to harder-than-average forecast horizons, while \(w_j<1\) correspond to easier horizons.

\subsection{Trajectory Difficulty Stratification}
\label{subsec:trajectory_stratification}

TRACE-CRC stratifies predicted CSI trajectories according to estimated prediction difficulty. The goal is to learn a simple mapping from a trajectory-level feature vector \(\mathbf{x}_i\), computed from the predicted trajectory, to a predicted difficulty score \(\widehat d_i\). The resulting scores are used to form lower- and higher-difficulty groups.

For each profile index \(i\in\mathcal D_{\mathrm{prof}}\), we define
\[
    z^{(i)}_j
    =
    \left\|
    \widehat{\mathbf H}^{(i)}_{t+j}
    \right\|_F,
    \qquad
    j=1,\dots,N_{\mathrm f},
\]
and construct the trajectory-level feature vector
\begin{equation}
\label{eq:x_i}
    \mathbf{x}_i
    =
    \left[
    \mu_i,\,
    \sigma_i,\,
    z^{(i)}_{\max},\,
    z^{(i)}_{\min},\,
    \mathrm{range}_i,\,
    \mathrm{slope}_i,\,
    \mathrm{TV}_i,\,
    \mathrm{curvature}_i,\,
    z^{(i)}_1,\,
    z^{(i)}_{N_{\mathrm f}}
    \right].
\end{equation}
Here, \(\mu_i\), \(\sigma_i\), \(z^{(i)}_{\max}\), and \(z^{(i)}_{\min}\) are the mean, standard deviation, maximum, and minimum of the sequence \(\{z^{(i)}_j\}_{j=1}^{N_{\mathrm f}}\), with \(\mathrm{range}_i=z^{(i)}_{\max}-z^{(i)}_{\min}\) and \(\mathrm{slope}_i=z^{(i)}_{N_{\mathrm f}}-z^{(i)}_1\). The term \(\mathrm{TV}_i\) denotes trajectory variation, defined as the mean absolute first-order difference of the predicted norm sequence, while \(\mathrm{curvature}_i\) denotes the mean absolute second-order difference:
\[
    \mathrm{TV}_i
    =
    \frac{1}{N_{\mathrm f}-1}
    \sum_{j=1}^{N_{\mathrm f}-1}
    \left|
    z^{(i)}_{j+1}
    -
    z^{(i)}_j
    \right|,
    \qquad
    \mathrm{curvature}_i
    =
    \frac{1}{N_{\mathrm f}-2}
    \sum_{j=2}^{N_{\mathrm f}-1}
    \left|
    z^{(i)}_{j+1}
    -
    2z^{(i)}_j
    +
    z^{(i)}_{j-1}
    \right|.
\]
These features summarize the magnitude, variability, trend, and local smoothness of the predicted CSI trajectory.

For the same profile index \(i\in\mathcal D_{\mathrm{prof}}\), where the true future CSI is available for calibration, we define the residual-based trajectory difficulty target as
\begin{equation}
\label{eq:traj_diff}
    d_i
    =
    \max_{1\le j\le N_{\mathrm f}}
    \frac{E_{i,j}}{w_j}.
\end{equation}
This score measures how difficult trajectory \(i\) is after accounting for the typical error scale at each forecast horizon. A large value of \(d_i\) indicates that the trajectory contains at least one unusually large residual relative to the expected horizon difficulty.

We then fit a ridge regression model
\(\widehat d(\mathbf{x})=\beta_0+\beta^\top \mathbf{x}\) on
\(\{(\mathbf x_i,d_i):i\in\mathcal D_{\mathrm{prof}}\}\), using regularization parameter \(\eta\).
We use ridge regression to keep the stratification model low-variance and stable under limited calibration data.
Afterwards, we apply it to the trajectories indexed by \(\mathcal D_{\mathrm{cp}}\). For each \(\ell\in\mathcal D_{\mathrm{cp}}\), this gives a predicted difficulty score
\(
    \widehat d_\ell = \widehat d(\mathbf{x}_\ell).
\)
For two strata, we set
\(\tau=\mathrm{median}\{\widehat d_\ell:\ell\in\mathcal D_{\mathrm{cp}}\}\)
and assign each conformal calibration trajectory by
\(g_\ell=\mathbf 1\{\widehat d_\ell>\tau\}\). 
Thus, \(g_\ell=0\) denotes the lower-difficulty group and \(g_\ell=1\) denotes the higher-difficulty group.
The same fitted model and threshold are used to assign any validation or test trajectory \(i\) to a group,
\begin{equation}
\label{eq:group_assignment_rule}
    g_i
    =
    \mathbf 1\{\widehat d(\mathbf x_i)>\tau\}.
\end{equation}
\subsection{Group-wise Conformal Calibration}
\label{sec:groupwise_calibration}

Given the horizon profile in~\eqref{eq:horizon_diff} and the group assignment rule in~\eqref{eq:group_assignment_rule}, TRACE-CRC calibrates a conformal scale within each difficulty stratum. For each conformal calibration index \(\ell\in\mathcal D_{\mathrm{cp}}\), where the true future CSI is available, we compute the same trajectory difficulty score defined in~\eqref{eq:traj_diff},
\[
    d_{\ell}
    =
    \max_{1\le j\le N_{\mathrm f}}
    \frac{E_{\ell,j}}{w_j}.
\]
On \(\mathcal D_{\mathrm{prof}}\), this score was used to fit the trajectory difficulty model; on \(\mathcal D_{\mathrm{cp}}\), it is used as the conformal calibration score.
For each group \(g\in\{0,1\}\), let
\[
    \mathcal S_g
    =
    \{d_\ell:\ell\in\mathcal D_{\mathrm{cp}},\, g_\ell=g\},
    \qquad
    n_g=|\mathcal S_g|.
\]
The group-wise conformal quantile \(q_g\) is
\[
    q_g
    =
    Q_{1-\alpha_{\mathrm{cp}}}^{+}(\mathcal S_g),
\]
where \(Q_{1-\alpha_{\mathrm{cp}}}^{+}\) denotes the split-conformal empirical quantile at level
\(\lceil(n_g+1)(1-\alpha_{\mathrm{cp}})\rceil/n_g\). If
\(\lceil(n_g+1)(1-\alpha_{\mathrm{cp}})\rceil>n_g\), we set \(q_g=+\infty\). Here, \(\alpha_{\mathrm{cp}}\) is the group-wise conformal calibration level. Thus, \(q_g\) is computed from the residual-based calibration scores of trajectories assigned to group \(g\).

For a generic group \(g\in\{0,1\}\), the preliminary horizon-dependent radius is
\begin{equation}
    r_{g,j}
    =
    q_g w_j,
    \qquad
    j=1,\dots,N_{\mathrm f}.
\end{equation}
Thus a trajectory \(i\) assigned to group \(g_i\) uses \(r_{g_i,j}=q_{g_i}w_j\).
Since \(d_\ell\le q_g\) implies \(E_{\ell,j}\le q_g w_j\) for every horizon \(j\), \(q_g\) controls the trajectory-level scale within each difficulty group, while \(w_j\) distributes that scale across forecast horizons.

\subsection{Learn-then-Test Risk Control}
\label{sec:ltt_risk_control}

The group-wise conformal calibration step produces preliminary radii \(q_g w_j\). TRACE-CRC then applies LTT risk control to select a global multiplier \(\lambda\) for these radii. Let \(\Lambda=\{\lambda_1,\dots,\lambda_m\}\) be a finite grid of candidate multipliers. For a validation index \(i\in\mathcal D_{\mathrm{val}}\), let \(g_i\in\{0,1\}\) denote its assigned difficulty group. For each \(\lambda\in\Lambda\), the candidate radius at horizon \(j\) is
\[
    r_{i,j}(\lambda)
    =
    r_{g_i,j}(\lambda)
    =
    \lambda q_{g_i} w_j .
\]

On the validation subset, we evaluate each candidate multiplier using the trajectory-level failure loss
\(
    L_i(\lambda)
=
\mathbf{1}
\left\{
\exists j\in\{1,\dots,N_{\mathrm f}\}:
E_{i,j}
>
\lambda q_{g_i} w_j
\right\}.
\)
Thus, \(L_i(\lambda)=1\) if at least one future CSI frame in trajectory \(i\) falls outside its corresponding Frobenius uncertainty ball. The empirical validation risk is
\[
    \widehat{\mathcal R}_{\mathrm{traj}}(\lambda)
    =
    \frac{1}{n_{\mathrm{val}}}
    \sum_{i\in\mathcal D_{\mathrm{val}}}
    L_i(\lambda),
    \qquad
    n_{\mathrm{val}}=|\mathcal D_{\mathrm{val}}|,
\]
and the corresponding population trajectory risk for a fresh CSI trajectory sample is
\(
    \mathcal{R}_{\mathrm{traj}}(\lambda)
    =
    \mathbb P\left(L_i(\lambda)=1\right).
\)
For each candidate multiplier, TRACE-CRC tests
\[
    H_0(\lambda):\mathcal{R}_{\mathrm{traj}}(\lambda)>\alpha
    \qquad
    \text{against}
    \qquad
    H_1(\lambda):\mathcal{R}_{\mathrm{traj}}(\lambda)\le\alpha,
\]
where \(\alpha\) is the target trajectory failure level. We test this hypothesis using a one-sided Hoeffding--Bentkus (HB) \(p\)-value
\[
    p(\lambda)
    =
    p_{\mathrm{HB}}
    \left(
    \widehat{\mathcal{R}}_{\mathrm{traj}}(\lambda),
    n_{\mathrm{val}},
    \alpha
    \right)
\]
\citep{Bentkus_2004}.
Let \(\mathcal A\subseteq\Lambda\) be the set of accepted multipliers after correction. TRACE-CRC selects
\(
    \lambda^\star
    =
    \min_{\lambda\in\mathcal A}\lambda,
\)
which gives the smallest certified multiplier among the accepted candidates. If \(\mathcal A=\emptyset\), no multiplier in the candidate grid is certified; in this case, the grid must be enlarged, or the method returns no certified radius on the grid.

For trajectory \(i\), the final radius at horizon \(j\) is
\begin{equation}
\label{eq:trace_final_radius}
r_{i,j}^{\star}
    =
    r_{g_i,j}^{\star}
    =
    \lambda^\star q_{g_i} w_j .    
\end{equation}

\subsection{Prediction Bands and Risk Control Guarantee}
\label{sec:prediction_bands_guarantee}

For a new CSI context, TRACE-CRC first computes the trajectory-level feature vector \(\mathbf x_i\) from the predicted future CSI trajectory, as defined in~\eqref{eq:x_i}. It then assigns the trajectory to a difficulty group \(g_i\) using~\eqref{eq:group_assignment_rule} and applies the final radius \(r_{i,j}^{\star}\) from~\eqref{eq:trace_final_radius} at each forecast horizon. The resulting trajectory-level prediction band is the collection of horizon-wise Frobenius balls
\[
    \widehat{\mathcal B}^{\star}_i
    =
    \left\{
    \mathcal B_{i,j}(r_{i,j}^{\star})
    \right\}_{j=1}^{N_{\mathrm f}},
\]
where each ball is centered at the corresponding predicted CSI frame \(\widehat{\mathbf H}^{(i)}_{t+j}\), as defined in~\eqref{eq:f_ball}.

Under trajectory-level exchangeability between the validation losses used by LTT and future deployment losses, and with FWER controlled at level \(\delta\), the selected rule \(\widehat{\mathcal B}^{\star}\) satisfies the CRC guarantee in~\eqref{eq:crc_goal}, certifying trajectory-level risk at level \(\alpha\). As discussed in Section~\ref{sec:dataset}, this assumption concerns exchangeability across complete trajectories and does not require independence among frames within a trajectory.

\section{Experimental Setup}
\label{sec:experimental_setup}

We evaluate TRACE-CRC on the multi-step CSI prediction task using the pretrained diffusion-based CSI predictor of 
\citet{sattari2025csipredictionusingdiffusion}; see
Sections~\ref{sec:dataset} and~\ref{subsec:csi_prediction} for details. The predictor was trained on \(10{,}000\) trajectory-level samples following the training procedure described in~\citet{sattari2025csipredictionusingdiffusion}. We use an inference SNR of \(20\,\mathrm{dB}\) and \(20\) reverse-diffusion sampling steps.
TRACE-CRC is applied post hoc to the pretrained CSI predictor. Its calibration is performed offline, while deployment requires only evaluation of the calibrated scaling rule and uncertainty radii. The online computational cost is therefore expected to be small relative to diffusion inference.
For the complex-valued CSI matrices, all residuals and uncertainty radii are computed using the Frobenius norm.
From each \(100\)-frame CSI sample described in Section~\ref{sec:dataset}, we extract a \(30\)-frame trajectory for prediction. The predictor uses \(N_p=10\) past frames as input and predicts \(N_f=20\) future frames. Thus, conformal calibration and evaluation are performed over a forecast horizon of \(N_f=20\).

At inference time, the \(1{,}000\) prediction trajectories are partitioned using a fixed random holdout. We assign 30\% of the trajectories to calibration and retain 70\% for testing, yielding \(n_{\mathrm{cal}}=300\) and \(n_{\mathrm{test}}=700\). This split is shared by all conformal methods and ensures that no CSI frames from the same trajectory are assigned to both calibration and test sets. We use a relatively small calibration set to evaluate TRACE-CRC in a limited-calibration regime while retaining a large independent test set for stable estimation of trajectory-level performance.
Since diffusion inference is stochastic, the complete evaluation is repeated over five random inference seeds while keeping the same calibration and test split, and results are aggregated across runs. For online adaptive methods, test trajectories are processed in a fixed dataset-index order for reproducibility; this order does not represent temporal progression.

The calibration trajectories are further divided into three disjoint subsets required by TRACE-CRC: the profile subset \(\mathcal D_{\mathrm{prof}}\), the conformal calibration subset \(\mathcal D_{\mathrm{cp}}\), and the validation subset \(\mathcal D_{\mathrm{val}}\). We allocate \(10\%\) of the calibration trajectories to profile learning, \(15\%\) of the remaining calibration trajectories to group-wise conformal calibration, and the rest to LTT validation, resulting in \(n_{\mathrm{prof}}=30\), \(n_{\mathrm{cp}}=40\), and \(n_{\mathrm{val}}=230\). 
We allocate the largest calibration split to LTT validation because the risk-control certificate depends directly on validation evidence.
We set \(\alpha=0.10\), corresponding to a target trajectory coverage of 0.90, and use \(\alpha_{\mathrm{cp}}=\alpha_h=0.10\) for group-wise conformal calibration and horizon-profile estimation.
For LTT, we set \(\delta=0.10\), corresponding to a \(90\%\) confidence level for the risk-control certificate. These values define a representative operating point and are kept fixed across methods to enable a controlled comparison of their calibration strategies. 
The LTT candidate family consists of \(27\) prespecified multipliers, including \(10\) uniformly spaced on \([0.70,1.30]\) and additional candidates on \([1.40,3.00]\). The \(\lambda\)-grid allows both mild deflation and inflation of the preliminary conformal radii. Additional method-specific parameters are reported in Appendix~\ref{app:method_parameters}.

\subsection{Baseline Methods}
\label{sec:baseline_methods}

We compare TRACE-CRC with two groups of methods: conformal baselines from the literature and ablations of the proposed conformal risk control framework. Since the methods from the literature do not all target the same formal guarantee, we organize them into four categories: standard split-conformal baselines, time-series adaptive conformal baselines, structured multi-step baselines, and risk control ablations. Within each category, when applicable, we distinguish horizon-wise variants from trajectory-level variants according to whether calibration targets individual forecast steps or the full CSI trajectory. All methods are evaluated on the same held-out test trajectories using the reliability and efficiency metrics defined in Section~\ref{sec:results_discussion}.

\paragraph{Standard split-conformal baselines.}
We first include standard split-conformal regression baselines \citep{papadopoulos2002inductive,lei2018distribution}. Global Residual Conformal pools all calibration residuals \(E_{i,j}\) across samples and horizons and uses a single radius \(R_{i,j}=q\). Horizon-wise conformal instead computes a separate quantile \(q_j\) for each forecast horizon, yielding \(R_{i,j}=q_j\). To represent trajectory-level split conformal calibration, we include Max-score split conformal, which uses the nonconformity score \(S_i=\max_{1\le j\le N_{\mathrm f}}E_{i,j}\) and applies the resulting quantile \(q_{\max}\) uniformly across all horizons. We also include a residual-quantile conformal baseline inspired by conformalized quantile regression~\citep{romano2019conformalized}. This baseline estimates a horizon-dependent residual quantile profile and then conformalizes it using the maximum residual excess over that profile.

\paragraph{Time-series adaptive conformal baselines.}
To compare against adaptive conformal methods designed for temporally ordered or nonstationary settings, we include exponentially weighted (EW) conformal variants.
EW-Horizon conformal and EW-Trajectory conformal replace the unweighted empirical quantile with a recency-weighted quantile, assigning larger weights to more recent residuals. These baselines are motivated by conformal prediction under distribution shift and non-exchangeability \citep{tibshirani2019conformal,barber2023conformal,wang2024online}.

We also include EnbPI \citep{xu2021conformal}, ACI \citep{gibbs2021adaptive}, and AgACI \citep{zaffran2022adaptive} as established adaptive online conformal time-series baselines. For each method, we report both a horizon and a trajectory variant. The horizon variant calibrates separately at each prediction step \(j\), while the trajectory variant first reduces each predicted CSI trajectory to a scalar score, typically \(\max_{1\le j\le N_f}E_{i,j}\), and then calibrates this trajectory-level score. This distinction is important because horizon variants primarily target per-horizon reliability, whereas trajectory variants are more directly aligned with the trajectory-level objective considered in this work.

\paragraph{Structured multi-step baselines.}
We include two CRC baselines, Bonferroni-CRC and Sidak-CRC, which use classical multi-step corrections. Both apply simultaneous inference corrections across forecast horizons, using the Bonferroni \citep{bonferroni1936teoria} and Sidak \citep{sidak1967rectangular} adjustments, respectively. Such corrections are commonly used to convert marginal intervals into simultaneous multi-step prediction intervals \citep{ravishanker1991multiple}. We also compare against CopulaCPTS, a joint multi-step conformal forecasting method that accounts for dependence across forecast horizons through an empirical copula construction \citep{sun2024copula}.

\paragraph{risk control ablations.}
Finally, we include ablations of the proposed risk control framework to isolate the contribution of each component. Table~\ref{tab:trace_crc_ablation_structure} summarizes which methods use horizon profiling \(w_j\), trajectory group-wise conformal quantiles \(q_g\), and LTT risk control calibration \citep{bates2021distributionfree,Angelopoulos:22CRC}. 
Global-CRC uses only the LTT multiplier \(\lambda^\star\). Horizon-Profile CRC adds \(w_j\), Trajectory-Stratified CRC adds \(q_g\), and TRACE-CRC combines all three components.
These ablations test whether LTT risk control alone is sufficient, or whether efficiency improves when risk control is combined with horizon profiling and trajectory stratification. 
 
\begin{table}[t]
\centering
\caption{Ablation structure for TRACE-CRC components.}
\label{tab:trace_crc_ablation_structure}
\begin{tabular}{lccc}
\toprule
\textbf{Method} & \textbf{\(w_j\)} & \textbf{\(q_g\)} & \textbf{\(\lambda^\star\)} \\
\midrule
Global-CRC                     & \xmark & \xmark & \cmark \\
Horizon-Profile CRC            & \cmark & \xmark & \cmark \\
Trajectory-Stratified CRC      & \xmark & \cmark & \cmark \\
TRACE-CRC                      & \cmark & \cmark & \cmark \\
\bottomrule
\end{tabular}
\end{table}

\section{Results and Discussion}
\label{sec:results_discussion}

We evaluate all methods on the same held-out CSI test trajectories. Our primary reliability metric is trajectory coverage (TC), and our efficiency metric is the horizon-wise average Frobenius radius \(\mathrm{AFR}_j\):
\[
    \mathrm{TC}
    =
    \frac{1}{n_{\mathrm{test}}}
    \sum_{i=1}^{n_{\mathrm{test}}}
    \mathbf 1
    \left\{
    \forall j\in\{1,\dots,N_{\mathrm f}\}:
    E_{i,j}\le r_{i,j}
    \right\},
    \qquad
    \mathrm{AFR}_j
    =
    \frac{1}{n_{\mathrm{test}}}
    \sum_{i=1}^{n_{\mathrm{test}}}
    r_{i,j}.
\]
We also report horizon-wise coverage \(\mathrm{HC}_j\) and summarize it using mean horizon coverage (MHC) and worst-horizon coverage (WHC):
\[
    \mathrm{HC}_j
    =
    \frac{1}{n_{\mathrm{test}}}
    \sum_{i=1}^{n_{\mathrm{test}}}
    \mathbf 1\{E_{i,j}\le r_{i,j}\},
    \qquad
    \mathrm{MHC}
    =
    \frac{1}{N_{\mathrm f}}\sum_{j=1}^{N_{\mathrm f}}\mathrm{HC}_j,
    \qquad
    \mathrm{WHC}
    =
    \min_{1\le j\le N_{\mathrm f}} \mathrm{HC}_j .
\]

Figures show the horizon-wise coverage \(\mathrm{HC}_j\) and average Frobenius radius \(\mathrm{AFR}_j\), averaged over five random diffusion inference seeds; shaded regions denote one standard deviation across seeds. For each seed, MHC is the mean of \(\mathrm{HC}_j\) across horizons, WHC is its minimum, and AFR is the mean of \(\mathrm{AFR}_j\). Tables report the mean \(\pm\) standard deviation of these per-seed metrics. MHC, WHC, and TC measure reliability relative to the target coverage \(1-\alpha=0.90\), while AFR measures efficiency.
Table~\ref{tab:best-methods-compact} presents representative methods from each comparison category; complete results and tuning parameters are reported in Appendices~\ref{app:complete_results} and~\ref{app:method_parameters}. 

Horizon-wise methods produce compact uncertainty regions but insufficient trajectory coverage, with TC values of \(0.710\) for Horizon-wise conformal and \(0.744\) for AgACI-Horizon. Methods that explicitly target full-trajectory reliability generally improve TC but require larger radii; for example, EnbPI-Trajectory and Bonferroni-CRC attain TC values of \(0.919\) and \(0.968\), with AFR values of \(21.80\) and \(20.00\), respectively. TRACE-CRC achieves \(\mathrm{TC}=0.933\pm0.011\) and \(\mathrm{AFR}=13.64\pm0.41\), exceeding the \(0.90\) target while remaining more efficient than the trajectory-level and simultaneous representatives. It also uses smaller radii than Residual Quantile and CopulaCPTS, while achieving higher trajectory coverage.

\begin{table}[t]
\centering
\small
\setlength{\tabcolsep}{4pt}
\renewcommand{\arraystretch}{1.12}
\caption{
Representative conformal methods for CSI uncertainty under the fixed random
trajectory-level split. The target trajectory coverage is \(0.90\).
Results are reported as mean \(\pm\) standard deviation across five diffusion
inference seeds. For each seed, MHC is the mean coverage across forecast
horizons, WHC is the worst-horizon coverage, TC is full-trajectory coverage,
and AFR is the mean Frobenius radius across forecast horizons and test
trajectories.
}
\label{tab:best-methods-compact}
\resizebox{\textwidth}{!}{%
\begin{tabular}{llcccc}
\toprule
Category & Method & MHC & WHC & TC & AFR \\
\midrule

\multirow{2}{*}{Standard split-conformal}
& Horizon-wise
& \(0.872 \pm 0.002\)
& \(0.855 \pm 0.006\)
& \(0.710 \pm 0.009\)
& \(10.73 \pm 0.06\) \\

& Residual Quantile
& \(0.965 \pm 0.002\)
& \(0.939 \pm 0.003\)
& \(0.903 \pm 0.007\)
& \(14.30 \pm 0.21\) \\

\midrule

\multirow{2}{*}{Time-series adaptive}
& AgACI-Horizon
& \(0.896 \pm 0.001\)
& \(0.891 \pm 0.001\)
& \(0.744 \pm 0.006\)
& \(11.49 \pm 0.06\) \\

& EnbPI-Trajectory
& \(0.984 \pm 0.001\)
& \(0.937 \pm 0.007\)
& \(0.919 \pm 0.008\)
& \(21.80 \pm 0.24\) \\

\midrule

\multirow{2}{*}{Structured multi-step}
& Bonferroni-CRC
& \(0.993 \pm 0.001\)
& \(0.985 \pm 0.003\)
& \(0.968 \pm 0.004\)
& \(20.00 \pm 0.29\) \\

& CopulaCPTS
& \(0.969 \pm 0.004\)
& \(0.945 \pm 0.004\)
& \(0.892 \pm 0.012\)
& \(14.87 \pm 0.29\) \\

\midrule

\multirow{2}{*}{Risk control / ablation}
& Global-CRC
& \(0.992 \pm 0.001\)
& \(0.967 \pm 0.005\)
& \(0.957 \pm 0.007\)
& \(24.37 \pm 0.54\) \\

& \textbf{TRACE-CRC}
& \(0.978 \pm 0.004\)
& \(0.962 \pm 0.006\)
& \(0.933 \pm 0.011\)
& \(13.64 \pm 0.41\) \\

\bottomrule
\end{tabular}%
}
\end{table}

\begin{figure}[t]
    \centering
    \input{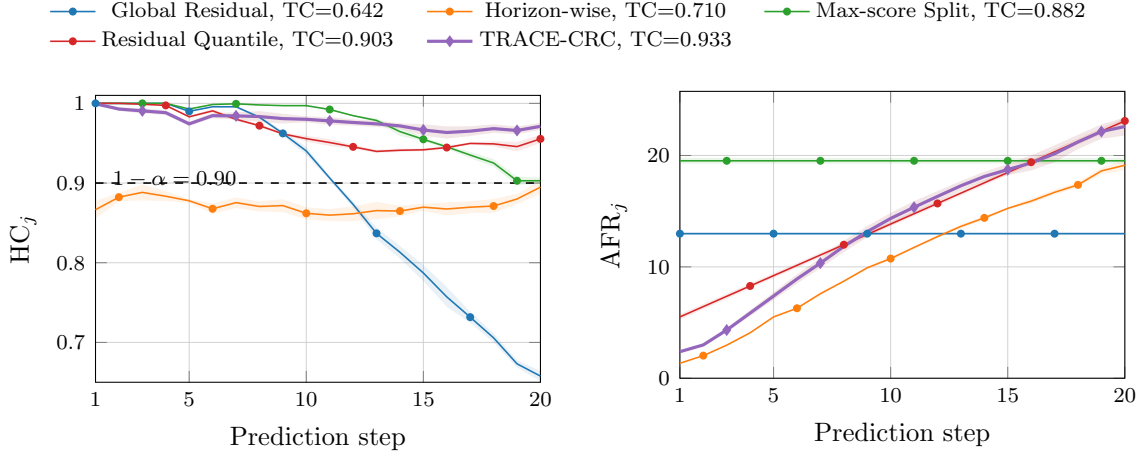}
    \caption{Horizon-wise coverage \(\mathrm{HC}_j\) (left) and average Frobenius radius \(\mathrm{AFR}_j\) (right) for standard split-conformal baselines and TRACE-CRC. Curves show means over five diffusion inference seeds, and shaded regions indicate \(\pm1\) standard deviation across seeds. The dashed line marks the target coverage \(1-\alpha=0.90\); legend entries report trajectory coverage (TC).}
    \label{fig:standard-split-conformal-coverage-radius}
\end{figure}

\paragraph{Standard split-conformal baselines.}
Figure~\ref{fig:standard-split-conformal-coverage-radius} compares the standard split-conformal baselines with TRACE-CRC. Global Residual Conformal provides high coverage at early horizons but degrades sharply over the forecast path, resulting in low trajectory coverage. Horizon-wise Conformal produces the most compact uncertainty balls, but its horizon-wise coverage remains mostly below the target and its TC is only \(0.710\). Residual Quantile Conformal better captures horizon-dependent error growth and reaches the trajectory target, with \(\mathrm{TC}=0.903\), although at a larger average radius than TRACE-CRC. Max-score Split Conformal directly calibrates a trajectory-level score, but produces substantially wider balls while still falling below the target, with \(\mathrm{TC}=0.882\). TRACE-CRC achieves the strongest reliability--efficiency trade-off in this group, attaining \(\mathrm{TC}=0.933\) with smaller radii than both trajectory-oriented split-conformal baselines.

\begin{figure}[t]
    \centering
    \input{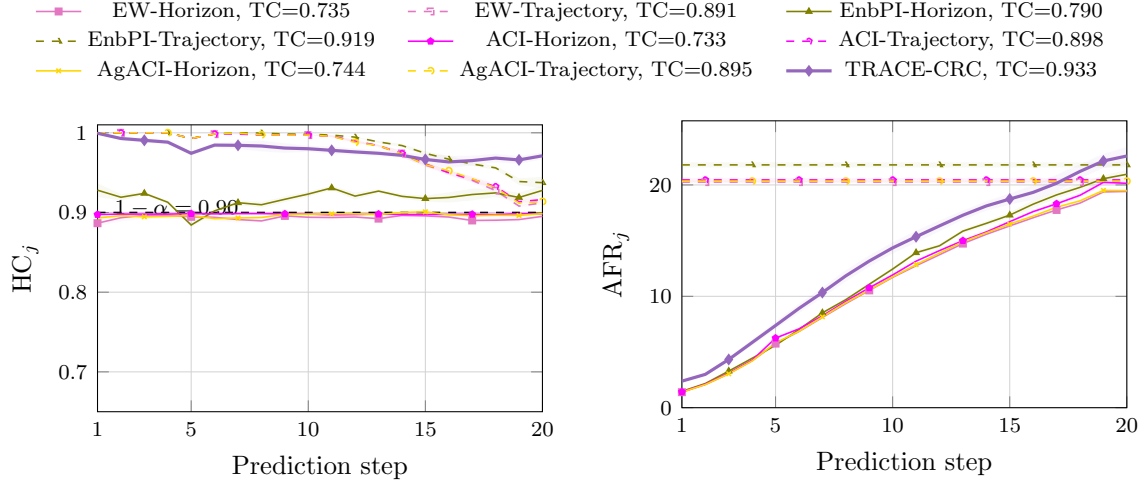}
    \caption{Horizon-wise coverage and average Frobenius radius for time-series adaptive conformal baselines and TRACE-CRC. Dashed curves denote trajectory-level variants, solid curves denote horizon-wise variants, and the diamond-marked curve denotes TRACE-CRC.}
    \label{fig:adaptive-coverage-radius}
\end{figure}

\paragraph{Time-series adaptive conformal baselines.}
Figure~\ref{fig:adaptive-coverage-radius} compares TRACE-CRC with adaptive time-series conformal baselines. Horizon-level adaptive methods remain relatively efficient, with AFR values between \(11.40\) and \(12.13\), but their trajectory coverage is well below the target, with TC ranging from \(0.733\) to \(0.790\). Trajectory-level variants improve full-path reliability, reaching TC values between \(0.891\) and \(0.919\), but their AFRs increase substantially to \(20.25\)--\(21.80\). Their horizon-wise coverage is high at early prediction steps but generally decreases toward later horizons, reflecting the increasing difficulty of long-range prediction. TRACE-CRC provides higher trajectory coverage than all adaptive baselines while remaining substantially more efficient than their trajectory-level variants. These results indicate that temporal adaptation alone is insufficient for efficient trajectory-level reliability.

\begin{figure}[t]
    \centering
    \input{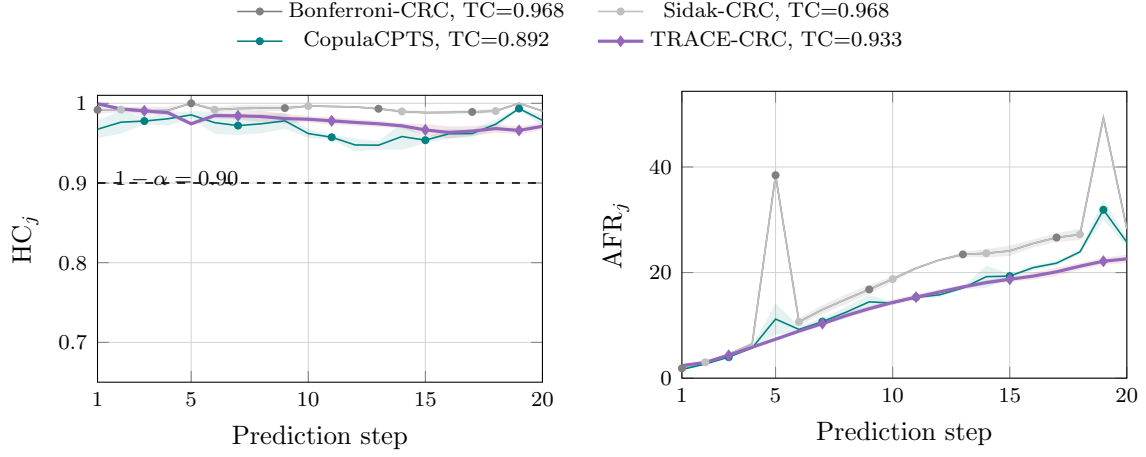}
\caption{Horizon-wise coverage and average Frobenius radius for structured multi-step baselines and TRACE-CRC.}
    \label{fig:structured-coverage-radius}
\end{figure}

\paragraph{Structured multi-step baselines.}
Figure~\ref{fig:structured-coverage-radius} compares TRACE-CRC with structured multi-step baselines. Bonferroni-CRC and Sidak-CRC achieve high trajectory coverage, with \(\mathrm{TC}=0.968\), but their simultaneous corrections are highly conservative, yielding an average AFR of \(20.00\) and sharp radius increases at several horizons. Their curves coincide because both corrections select the same split-conformal order statistic in this finite-sample setting. CopulaCPTS is less conservative, with \(\mathrm{AFR}=14.87\), but remains below the trajectory target at \(\mathrm{TC}=0.892\); its horizon-wise coverage also exhibits greater variability across diffusion inference seeds at several prediction steps. TRACE-CRC attains \(\mathrm{TC}=0.933\) with the smallest average radius in this group, \(\mathrm{AFR}=13.64\), providing a more favorable reliability--efficiency trade-off than the structured baselines.

\begin{figure}[t]
    \centering
    \input{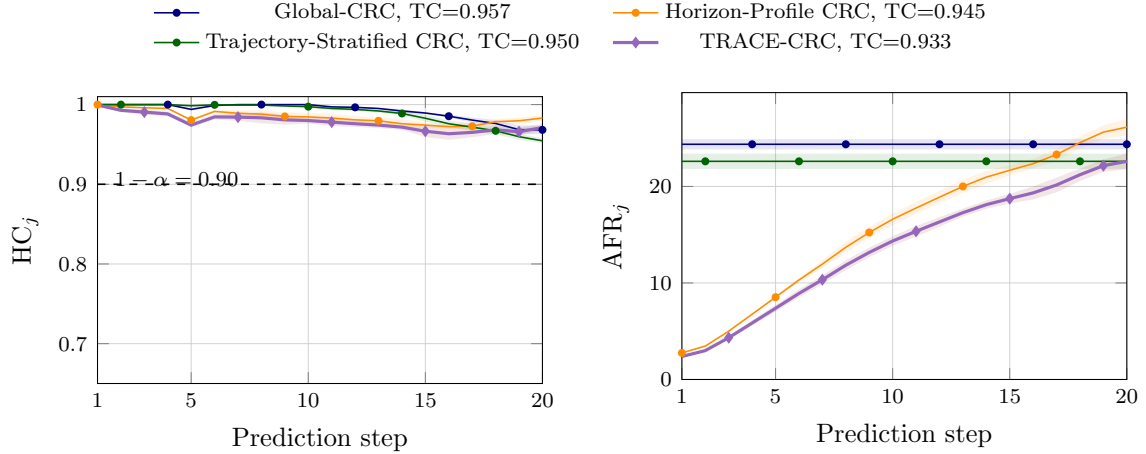}
    \caption{Horizon-wise coverage and average Frobenius radius for TRACE-CRC and its risk-control ablations.}
    \label{fig:risk control-coverage-radius}
\end{figure}

\paragraph{Risk control ablations.}
Figure~\ref{fig:risk control-coverage-radius} examines how horizon profiling and trajectory stratification affect the reliability--efficiency trade-off within the risk-control framework. Global-CRC is the most conservative variant, attaining \(\mathrm{TC}=0.957\) with a constant \(\mathrm{AFR}=24.37\) across horizons. Trajectory-Stratified CRC achieves similar trajectory coverage, \(\mathrm{TC}=0.950\), while reducing the constant radius to \(22.60\). Horizon-Profile CRC yields horizon-dependent radii and lowers the average AFR to \(15.78\), with \(\mathrm{TC}=0.945\). TRACE-CRC combines both adaptations and achieves \(\mathrm{TC}=0.933\) with the smallest average radius, \(\mathrm{AFR}=13.64\). These results show that horizon profiling contributes most of the efficiency gain, while trajectory stratification provides an additional reduction in uncertainty radii when combined with horizon-aware calibration in TRACE-CRC.

\paragraph{Robustness to calibration partition.}
We evaluated TRACE-CRC over \(50\) runs using a fixed outer \(300/700\) calibration/test split while varying the internal allocation of the \(300\) calibration trajectories into \(\mathcal D_{\mathrm{prof}}\), \(\mathcal D_{\mathrm{cp}}\), and \(\mathcal D_{\mathrm{val}}\), together with five diffusion inference seeds. TRACE-CRC selected a formally certified multiplier in every run and achieved an average trajectory coverage of \(0.940\pm0.015\), exceeding the target level in every run. Variability was small and was driven primarily by the internal calibration partition rather than diffusion inference (Appendix~\ref{app:robustness}), indicating robustness to the particular calibration allocation.

\paragraph{Overall reliability--efficiency trade-off.}
Overall, MHC alone is not sufficient for evaluating multi-step CSI uncertainty: several methods achieve high MHC while their TC remains below the nominal target. Methods that explicitly target trajectory-level or simultaneous coverage improve TC but often increase AFR. TRACE-CRC provides a favorable trade-off by maintaining TC above the nominal target while reducing AFR relative to conservative simultaneous-correction baselines.

\section{Conclusion}
\label{sec:conclusion}

This work introduced TRACE-CRC, a trajectory-adaptive conformal risk control (CRC) framework for uncertainty quantification in multi-step CSI prediction. The central motivation is that standard marginal or horizon-wise coverage objectives are not sufficient when downstream decisions depend on the full predicted channel trajectory. TRACE-CRC constructs per-horizon Frobenius-norm uncertainty balls and calibrates them under a trajectory-level failure criterion, while incorporating horizon profiling and trajectory stratification to account for structured variation in CSI prediction errors.

Empirically, TRACE-CRC achieved a favorable reliability--efficiency trade-off. Horizon-wise and adaptive conformal baselines often produced compact uncertainty balls but failed to maintain reliable trajectory coverage, whereas conservative multi-step methods such as Bonferroni and Global-CRC improved trajectory coverage at the cost of substantially larger radii. The ablation study showed that combining horizon profiling with trajectory stratification reduces conservatism while maintaining trajectory-level reliability.
More broadly, TRACE-CRC could serve as a foundation for trajectory-level uncertainty quantification in other complex time-series use cases in telecommunications.
It may also have applications in areas that require reliability over an entire predicted sequence, including financial transaction monitoring, industrial sensing, and healthcare forecasting.

A limitation of the current study is that the risk-control certificate and the tightness of the resulting Frobenius-norm uncertainty balls depend on the available calibration and validation data, the forecast horizon length, and the assumption of trajectory-level exchangeability between validation and deployment. This assumption may fail when the distribution of complete trajectories changes, for example because SNR, mobility, or blockage regimes shift after calibration, in which case validation losses may no longer represent deployment losses. Future work should examine sensitivity to calibration-set size, validation-set size, prediction horizon length, and other distribution shifts inherent in wireless communications, in addition to shift-aware recalibration strategies.

\section*{Acknowledgment}
This work was supported in part by the EUREKA CELTIC-NEXT SUSTAINET-Advance project, funded by Vinnova (Sweden's Innovation Agency) under Grant 2025-02987, in part by the Swedish Research Council (VR) through the 6G-NTN-E Research Environment under Grant 2024-06645.

\bibliography{Refs}

@book{vovk/1062391,
author = {Vovk, Vladimir and Gammerman, Alex and Shafer, Glenn},
title = {Algorithmic Learning in a Random World},
year = {2005},
isbn = {0387001522},
publisher = {Springer-Verlag},
address = {Berlin, Heidelberg}
}

@misc{angelopoulos2022gentleintroductionconformalprediction,
  title={A Gentle Introduction to Conformal Prediction and Distribution-Free Uncertainty Quantification},
  author={Anastasios N. Angelopoulos and Stephen Bates},
  year={2022},
  eprint={2107.07511},
  archivePrefix={arXiv},
  primaryClass={cs.LG},
  url={https://arxiv.org/abs/2107.07511}
}

@misc{angelopoulos2022learntestcalibratingpredictive,
      title={Learn then Test: Calibrating Predictive Algorithms to Achieve Risk Control}, 
      author={Anastasios N. Angelopoulos and Stephen Bates and Emmanuel J. Candès and Michael I. Jordan and Lihua Lei},
      year={2022},
      eprint={2110.01052},
      archivePrefix={arXiv},
      primaryClass={cs.LG},
      url={https://arxiv.org/abs/2110.01052}, 
}

@article{bates2021distributionfree,
  title   = {Distribution-Free, Risk-Controlling Prediction Sets},
  author  = {Bates, Stephen and Angelopoulos, Anastasios and Lei, Lihua and Malik, Jitendra and Jordan, Michael I.},
  journal = {Journal of the ACM},
  volume  = {68},
  number  = {6},
  pages   = {43:1--43:34},
  year    = {2021},
  doi     = {10.1145/3478535},
  url     = {https://doi.org/10.1145/3478535}
}

@misc{Angelopoulos:22CRC,
  title={Conformal Risk Control},
  author={Angelopoulos, Anastasios N. and Bates, Stephen and Fisch, Adam and Lei, Lihua and Schuster, Tal},
  year={2022},
  eprint={2208.02814},
  archivePrefix={arXiv},
  primaryClass={stat.ME},
  doi={10.48550/arXiv.2208.02814},
  url={https://arxiv.org/abs/2208.02814}
}

@inproceedings{farinhas2024nonexchangeable,
  title     = {Non-Exchangeable Conformal Risk Control},
  author    = {Farinhas, Ant{\'o}nio F. and Ulmer, Dennis and Zerva, Chrysoula and Martins, Andr{\'e} F. T.},
  booktitle = {International Conference on Learning Representations},
  year      = {2024},
  url       = {https://openreview.net/forum?id=j511LaqEeP}
}

@inproceedings{zecchin2024localized,
  title     = {Localized Adaptive Risk Control},
  author    = {Zecchin, Matteo and Simeone, Osvaldo},
  booktitle = {Advances in Neural Information Processing Systems},
  volume    = {37},
  pages     = {8165--8192},
  year      = {2024},
  url       = {https://proceedings.neurips.cc/paper_files/paper/2024/hash/0f93c3e9b557980d93016671acd94bd2-Abstract-Conference.html}
}

@inproceedings{zhang2024fair,
  title     = {Fair Risk Control: A Generalized Framework for Calibrating Multi-group Fairness Risks},
  author    = {Zhang, Lujing and Roth, Aaron and Zhang, Linjun},
  booktitle = {Proceedings of the 41st International Conference on Machine Learning},
  year      = {2024},
  url       = {https://arxiv.org/abs/2405.02225}
}

@inproceedings{overman2024aligning,
  title     = {Aligning Model Properties via Conformal Risk Control},
  author    = {Overman, William and Vallon, Jacqueline Jil and Bayati, Mohsen},
  booktitle = {Advances in Neural Information Processing Systems},
  volume    = {37},
  pages     = {110702--110722},
  year      = {2024},
  url       = {https://proceedings.neurips.cc/paper_files/paper/2024/hash/c79625091a4f8b5d3abe29f3b14fa43a-Abstract-Conference.html}
}

@inproceedings{jazbec2024fast,
  title     = {Fast yet Safe: Early-Exiting with Risk Control},
  author    = {Jazbec, Metod and Timans, Alexander and Veljkovi{\'c}, Tin Had{\v z}i and Sakmann, Kaspar and Zhang, Dan and Naesseth, Christian A. and Nalisnick, Eric},
  booktitle = {Advances in Neural Information Processing Systems},
  volume    = {37},
  pages     = {129825--129854},
  year      = {2024},
  url       = {https://proceedings.neurips.cc/paper_files/paper/2024/hash/ea5a63f7ddb82e58623693fd1f4933f7-Abstract-Conference.html}
}

@misc{zecchin2024forking,
  title         = {Forking Uncertainties: Reliable Prediction and Model Predictive Control with Sequence Models via Conformal Risk Control},
  author        = {Zecchin, Matteo and Park, Sangwoo and Simeone, Osvaldo},
  year          = {2024},
  eprint        = {2310.10299},
  archivePrefix = {arXiv},
  primaryClass  = {eess.SY},
  url           = {https://arxiv.org/abs/2310.10299}
}

@article{Bentkus_2004,
   title={On Hoeffding’s inequalities},
   volume={32},
   ISSN={0091-1798},
   url={http://dx.doi.org/10.1214/009117904000000360},
   DOI={10.1214/009117904000000360},
   number={2},
   journal={The Annals of Probability},
   publisher={Institute of Mathematical Statistics},
   author={Bentkus, Vidmantas},
   year={2004},
   month=apr }

@book{bonferroni1936teoria,
  title={Teoria statistica delle classi e calcolo delle probabilit{\`a}},
  author={Bonferroni, Carlo Emilio},
  year={1936},
  publisher={Pubblicazioni del R Istituto Superiore di Scienze Economiche e Commerciali di Firenze}
}

@article{sidak1967rectangular,
  title={Rectangular Confidence Regions for the Means of Multivariate Normal Distributions},
  author={{\v{S}}id{\'a}k, Zbyn{\v{e}}k},
  journal={Journal of the American Statistical Association},
  volume={62},
  number={318},
  pages={626--633},
  year={1967},
  publisher={Taylor \& Francis}
}

@article{ravishanker1991multiple,
  title={Multiple prediction intervals for time series: Comparison of simultaneous and marginal intervals},
  author={Ravishanker, Nalini and Dey, Dipak K. and Tiao, George C.},
  journal={International Journal of Forecasting},
  volume={7},
  number={4},
  pages={445--463},
  year={1991},
  publisher={Elsevier}
}

@inproceedings{papadopoulos2002inductive,
  title        = {Inductive Confidence Machines for Regression},
  author       = {Papadopoulos, Harris and Proedrou, Kostas and Vovk, Volodya and Gammerman, Alex},
  booktitle    = {Machine Learning: ECML 2002},
  pages        = {345--356},
  year         = {2002},
  publisher    = {Springer},
  doi          = {10.1007/3-540-36755-1_29}
}

@article{lei2018distribution,
  title        = {Distribution-Free Predictive Inference for Regression},
  author       = {Lei, Jing and G'Sell, Max and Rinaldo, Alessandro and Tibshirani, Ryan J. and Wasserman, Larry},
  journal      = {Journal of the American Statistical Association},
  volume       = {113},
  number       = {523},
  pages        = {1094--1111},
  year         = {2018},
  publisher    = {Taylor \& Francis},
  doi          = {10.1080/01621459.2017.1307116},
  url          = {https://www.tandfonline.com/doi/abs/10.1080/01621459.2017.1307116}
}

@inproceedings{romano2019conformalized,
  title        = {Conformalized Quantile Regression},
  author       = {Romano, Yaniv and Patterson, Evan and Cand{\`e}s, Emmanuel J.},
  booktitle    = {Advances in Neural Information Processing Systems},
  volume       = {32},
  year         = {2019},
  url          = {https://papers.neurips.cc/paper_files/paper/2019/hash/5103c3584b063c431bd1268e9b5e76fb-Abstract.html}
}

@inproceedings{tibshirani2019conformal,
  title        = {Conformal Prediction Under Covariate Shift},
  author       = {Tibshirani, Ryan J. and Barber, Rina Foygel and Cand{\`e}s, Emmanuel J. and Ramdas, Aaditya},
  booktitle    = {Advances in Neural Information Processing Systems},
  volume       = {32},
  year         = {2019},
  url          = {https://arxiv.org/abs/1904.06019}
}

@article{barber2023conformal,
  title        = {Conformal Prediction Beyond Exchangeability},
  author       = {Barber, Rina Foygel and Cand{\`e}s, Emmanuel J. and Ramdas, Aaditya and Tibshirani, Ryan J.},
  journal      = {The Annals of Statistics},
  volume       = {51},
  number       = {2},
  pages        = {816--845},
  year         = {2023},
  publisher    = {Institute of Mathematical Statistics},
  doi          = {10.1214/23-AOS2276},
  url          = {https://projecteuclid.org/journals/annals-of-statistics/volume-51/issue-2/Conformal-prediction-beyond-exchangeability/10.1214/23-AOS2276.full}
}

@inproceedings{xu2021conformal,
  title        = {Conformal Prediction Interval for Dynamic Time-Series},
  author       = {Xu, Chen and Xie, Yao},
  booktitle    = {Proceedings of the 38th International Conference on Machine Learning},
  series       = {Proceedings of Machine Learning Research},
  volume       = {139},
  pages        = {11559--11569},
  year         = {2021},
  publisher    = {PMLR},
  url          = {https://proceedings.mlr.press/v139/xu21h.html}
}

@article{xu2023conformal,
  title        = {Conformal Prediction for Time Series},
  author       = {Xu, Chen and Xie, Yao},
  journal      = {IEEE Transactions on Pattern Analysis and Machine Intelligence},
  volume       = {45},
  number       = {10},
  pages        = {11575--11587},
  year         = {2023},
  publisher    = {IEEE},
  doi          = {10.1109/TPAMI.2023.3272339},
  url          = {https://ieeexplore.ieee.org/document/10121511}
}

@inproceedings{gibbs2021adaptive,
  title     = {Adaptive Conformal Inference Under Distribution Shift},
  author    = {Gibbs, Isaac and Cand{\`e}s, Emmanuel},
  booktitle = {Advances in Neural Information Processing Systems},
  volume    = {34},
  pages     = {1660--1672},
  year      = {2021},
  url       = {https://proceedings.neurips.cc/paper/2021/hash/0d441de75945e5acbc865406fc9a2559-Abstract.html}
}

@inproceedings{zaffran2022adaptive,
  title        = {Adaptive Conformal Predictions for Time Series},
  author       = {Zaffran, Margaux and Dieuleveut, Aymeric and F{\'e}ron, Olivier and Goude, Yannig and Josse, Julie},
  booktitle    = {Proceedings of the 39th International Conference on Machine Learning},
  series       = {Proceedings of Machine Learning Research},
  volume       = {162},
  pages        = {25834--25866},
  year         = {2022},
  publisher    = {PMLR},
  url          = {https://proceedings.mlr.press/v162/zaffran22a.html}
}

@inproceedings{xu2023sequential,
  title        = {Sequential Predictive Conformal Inference for Time Series},
  author       = {Xu, Chen and Xie, Yao},
  booktitle    = {Proceedings of the 40th International Conference on Machine Learning},
  series       = {Proceedings of Machine Learning Research},
  volume       = {202},
  pages        = {38707--38727},
  year         = {2023},
  publisher    = {PMLR},
  url          = {https://proceedings.mlr.press/v202/xu23r.html}
}

@inproceedings{lee2025kernel,
  title        = {Kernel-Based Optimally Weighted Conformal Time-Series Prediction},
  author       = {Lee, Jonghyeok and Xu, Chen and Xie, Yao},
  booktitle    = {International Conference on Learning Representations},
  year         = {2025},
  url          = {https://openreview.net/forum?id=oP7arLOWix}
}

@inproceedings{sun2024copula,
  title        = {Copula Conformal Prediction for Multi-step Time Series Forecasting},
  author       = {Sun, Sophia Huiwen and Yu, Rose},
  booktitle    = {International Conference on Learning Representations},
  year         = {2024},
  url          = {https://openreview.net/forum?id=ojIJZDNIBj}
}

@inproceedings{galvaolopes2024conforme,
  title        = {{ConForME}: Multi-horizon Conditional Conformal Time Series Forecasting},
  author       = {Galv{\~a}o Lopes, Aloysio and Goubault, Eric and Putot, Sylvie and Pautet, Laurent},
  booktitle    = {Proceedings of the Thirteenth Symposium on Conformal and Probabilistic Prediction with Applications},
  series       = {Proceedings of Machine Learning Research},
  volume       = {230},
  pages        = {345--365},
  year         = {2024},
  publisher    = {PMLR},
  url          = {https://proceedings.mlr.press/v230/galvao-lopes24a.html}
}

@misc{wang2024online,
  title        = {Online Conformal Inference for Multi-step Time Series Forecasting},
  author       = {Wang, Xiaoqian and Hyndman, Rob J.},
  year         = {2024},
  eprint       = {2410.13115},
  archivePrefix = {arXiv},
  primaryClass = {stat.ME},
  url          = {https://arxiv.org/abs/2410.13115}
}

@article{massivemimobook,
url = {http://dx.doi.org/10.1561/2000000093},
year = {2017},
volume = {11},
journal = {Foundations and Trends{\textregistered} in Signal Processing},
title = {Massive {MIMO} Networks: {Spectral}, Energy, and Hardware Efficiency},
doi = {10.1561/2000000093},
issn = {1932-8346},
number = {3-4},
pages = {154-655},
author = {Emil Bj\"{o}rnson and Jakob Hoydis and Luca Sanguinetti}
}

@ARTICLE{10422880,
  author={Stenhammar, O. and Fodor, G. and Fischione, C.},
  journal={IEEE Wireless Commun.}, 
  title={A comparison of neural networks for wireless channel prediction}, 
  year={2024},
  volume={31},
  number={3},
  pages={235--241},
  doi={10.1109/MWC.006.2300140}
}

@ARTICLE{9210016,
  author={Kim, H. and Kim, S. and Lee, H. and Jang, C. and Choi, Y. and Choi, J.},
  journal={IEEE Trans. Commun.}, 
  title={Massive {MIMO} channel prediction: {Kalman} filtering vs. machine learning}, 
  year={2021},
  volume={69},
  number={1},
  pages={518--528},
  doi={10.1109/TCOMM.2020.3027882}
}

@misc{sattari2025csipredictionusingdiffusion,
      title={CSI Prediction Using Diffusion Models}, 
      author={Mehdi Sattari and Javad Aliakbari and Alexandre Graell i Amat and Tommy Svensson},
      year={2025},
      eprint={2510.11214},
      archivePrefix={arXiv},
      primaryClass={eess.SP},
      url={https://arxiv.org/abs/2510.11214}, 
}

@ARTICLE{1512123,
  author={Baddour, K. E. and Beaulieu, N. C.},
  journal={IEEE Trans. Wireless Commun.}, 
  title={Autoregressive modeling for fading channel simulation}, 
  year={2005},
  volume={4},
  number={4},
  pages={1650--1662},
  doi={10.1109/TWC.2005.850327}
}

@ARTICLE{9127447,
  author={Yin, H. and Wang, H. and Liu, Y. and Gesbert, D.},
  journal={IEEE J. Sel. Areas Commun.}, 
  title={Addressing the curse of mobility in massive {MIMO} with Prony-based angular-delay domain channel predictions}, 
  year={2020},
  volume={38},
  number={12},
  pages={2903--2917},
  doi={10.1109/JSAC.2020.3005473}
}

@INPROCEEDINGS{8746352,
  author={Jiang, W. and Schotten, H. D.},
  booktitle={Proc. IEEE Veh. Technol. Conf. (VTC Spring)}, 
  title={Recurrent neural network-based frequency-domain channel prediction for wideband communications}, 
  year={2019},
  pages={1--6},
  doi={10.1109/VTCSpring.2019.8746352}
}

@ARTICLE{10965849,
  author={Jin, Y. and Wu, Y. and Gao, Y. and Zhang, S. and Xu, S. and Wang, C.-X.},
  journal={IEEE Trans. Wireless Commun.}, 
  title={LinFormer: A linear-based lightweight transformer architecture for time-aware {MIMO} channel prediction}, 
  year={2025},
  pages={1--1},
  doi={10.1109/TWC.2025.3558950}
}

@ARTICLE{11329107,
  author={Yoo, Seonghoon and Park, Sangwoo and Popovski, Petar and Kang, Joonhyuk and Simeone, Osvaldo},
  journal={IEEE Transactions on Signal Processing}, 
  title={Calibrating Wireless AI via Meta-Learned Context-Dependent Conformal Prediction}, 
  year={2026},
  volume={74},
  number={},
  pages={423-438},
  doi={10.1109/TSP.2026.3650912}}

@INPROCEEDINGS{11143438,
  author={Su, Xin and Hou, Qiushuo and He, Ruisi and Simeone, Osvaldo},
  booktitle={2025 IEEE 26th International Workshop on Signal Processing and Artificial Intelligence for Wireless Communications (SPAWC)}, 
  title={Conformal Robust Beamforming Via Generative Channel Models}, 
  year={2025},
  volume={},
  number={},
  pages={1-5},
  doi={10.1109/SPAWC66079.2025.11143438}}

@misc{kim2026mimochannelpredictiondeep,
      title={MIMO Channel Prediction via Deep Learning-based Conformal Bayes Filter}, 
      author={Dongwon Kim and Jinu Gong and Joonhyuk Kang},
      year={2026},
      eprint={2603.04764},
      archivePrefix={arXiv},
      primaryClass={eess.SP},
      url={https://arxiv.org/abs/2603.04764}, 
}

@misc{simeone2025conformalcalibrationensuringreliability,
      title={Conformal Calibration: Ensuring the Reliability of Black-Box AI in Wireless Systems}, 
      author={Osvaldo Simeone and Sangwoo Park and Matteo Zecchin},
      year={2025},
      eprint={2504.09310},
      archivePrefix={arXiv},
      primaryClass={cs.IT},
      url={https://arxiv.org/abs/2504.09310}, 
}

@INPROCEEDINGS{10096780,
  author={Cohen, Kfir M. and Park, Sangwoo and Simeone, Osvaldo and Shamai Shitz, Shlomo},
  booktitle={ICASSP 2023 - 2023 IEEE International Conference on Acoustics, Speech and Signal Processing (ICASSP)}, 
  title={Calibrating AI Models for Few-Shot Demodulation VIA Conformal Prediction}, 
  year={2023},
  volume={},
  number={},
  pages={1-5},
  doi={10.1109/ICASSP49357.2023.10096780}}

@ARTICLE{10262367,
  author={Cohen, Kfir M. and Park, Sangwoo and Simeone, Osvaldo and Shamai Shitz, Shlomo},
  journal={IEEE Transactions on Machine Learning in Communications and Networking}, 
  title={Calibrating AI Models for Wireless Communications via Conformal Prediction}, 
  year={2023},
  volume={1},
  number={},
  pages={296-312},
  doi={10.1109/TMLCN.2023.3319282}}

@TECHREPORT{3gpp.38.901,
  organization={3rd Generation Partnership Project (3GPP)},
  title={{Study on channel model for frequencies from 0.5 to 100 {GHz}}}, 
  institution={3GPP},
  number={TR 38.901},
  version={17.0.0},
  year={2022},
  month={Mar.},
  note={Available: \url{https://www.3gpp.org/ftp/Specs/archive/38_series/38.901/}}
}
\newpage
\appendix

\section{Complete Results}
\label{app:complete_results}

Table~\ref{tab:method-comparison-compact} reports the full set of evaluated conformal methods. 
This table includes all standard split-conformal baselines, adaptive time-series baselines, structured multi-step baselines, and risk control ablations.

\begin{table}[H]
\centering
\scriptsize
\setlength{\tabcolsep}{3pt}
\renewcommand{\arraystretch}{1.12}
\caption{Validity and efficiency of all conformal methods under the fixed \(300/700\) calibration--test split. Results are reported as mean \(\pm\) standard deviation over five random diffusion inference seeds; the target trajectory coverage is \(0.90\).}
\label{tab:method-comparison-compact}
\resizebox{\textwidth}{!}{%
\begin{tabular}{llcccc}
\toprule
Category & Method & MHC & WHC & TC & AFR \\
\midrule

\multirow{4}{*}{Standard split-conformal}
& Global Residual
& \(0.880 \pm 0.003\)
& \(0.658 \pm 0.006\)
& \(0.642 \pm 0.006\)
& \(12.98 \pm 0.13\) \\

& Horizon-wise
& \(0.872 \pm 0.002\)
& \(0.855 \pm 0.006\)
& \(0.710 \pm 0.009\)
& \(10.73 \pm 0.06\) \\

& Max-score Split
& \(0.973 \pm 0.001\)
& \(0.901 \pm 0.005\)
& \(0.882 \pm 0.005\)
& \(19.52 \pm 0.26\) \\

& Residual Quantile
& \(0.965 \pm 0.002\)
& \(0.939 \pm 0.003\)
& \(0.903 \pm 0.007\)
& \(14.30 \pm 0.21\) \\

\midrule

\multirow{8}{*}{Time-series adaptive}
& EW-Horizon
& \(0.893 \pm 0.001\)
& \(0.885 \pm 0.001\)
& \(0.735 \pm 0.005\)
& \(11.40 \pm 0.03\) \\

& EW-Trajectory
& \(0.976 \pm 0.001\)
& \(0.908 \pm 0.001\)
& \(0.891 \pm 0.002\)
& \(20.25 \pm 0.11\) \\

& EnbPI-Horizon
& \(0.918 \pm 0.003\)
& \(0.884 \pm 0.003\)
& \(0.790 \pm 0.010\)
& \(12.13 \pm 0.18\) \\

& EnbPI-Trajectory
& \(0.984 \pm 0.001\)
& \(0.937 \pm 0.007\)
& \(0.919 \pm 0.008\)
& \(21.80 \pm 0.24\) \\

& ACI-Horizon
& \(0.898 \pm 0.000\)
& \(0.896 \pm 0.001\)
& \(0.733 \pm 0.005\)
& \(11.74 \pm 0.03\) \\

& ACI-Trajectory
& \(0.977 \pm 0.001\)
& \(0.912 \pm 0.001\)
& \(0.898 \pm 0.001\)
& \(20.46 \pm 0.17\) \\

& AgACI-Horizon
& \(0.896 \pm 0.001\)
& \(0.891 \pm 0.001\)
& \(0.744 \pm 0.006\)
& \(11.49 \pm 0.06\) \\

& AgACI-Trajectory
& \(0.977 \pm 0.001\)
& \(0.911 \pm 0.003\)
& \(0.895 \pm 0.003\)
& \(20.34 \pm 0.15\) \\

\midrule

\multirow{3}{*}{Structured multi-step}
& Bonferroni-CRC
& \(0.993 \pm 0.001\)
& \(0.985 \pm 0.003\)
& \(0.968 \pm 0.004\)
& \(20.00 \pm 0.29\) \\

& Sidak-CRC
& \(0.993 \pm 0.001\)
& \(0.985 \pm 0.003\)
& \(0.968 \pm 0.004\)
& \(20.00 \pm 0.29\) \\

& CopulaCPTS
& \(0.969 \pm 0.004\)
& \(0.945 \pm 0.004\)
& \(0.892 \pm 0.012\)
& \(14.87 \pm 0.29\) \\

\midrule

\multirow{4}{*}{Risk control / ablation}
& Global-CRC
& \(0.992 \pm 0.001\)
& \(0.967 \pm 0.005\)
& \(0.957 \pm 0.007\)
& \(24.37 \pm 0.54\) \\

& Horizon-Profile CRC
& \(0.984 \pm 0.002\)
& \(0.970 \pm 0.004\)
& \(0.945 \pm 0.004\)
& \(15.78 \pm 0.38\) \\

& Trajectory-Stratified CRC
& \(0.989 \pm 0.002\)
& \(0.954 \pm 0.003\)
& \(0.950 \pm 0.005\)
& \(22.60 \pm 0.79\) \\

& \textbf{TRACE-CRC}
& \(0.978 \pm 0.004\)
& \(0.962 \pm 0.006\)
& \(0.933 \pm 0.011\)
& \(13.64 \pm 0.41\) \\

\bottomrule
\end{tabular}%
}
\end{table}

\FloatBarrier
\section{Method Parameters}
\label{app:method_parameters}

Tables~\ref{tab:baseline-selected-parameters} and
\ref{tab:baseline-validation-performance} report the selected baseline configurations and their validation performance, respectively. The search grids were fixed before evaluation. For each of the five random diffusion inference seeds used in the main experiment, the fixed \(300\)-trajectory calibration set was divided into \(210\) tuning-training and \(90\) tuning-validation trajectories; the \(700\) test trajectories were not used for selection.

Horizon-wise and trajectory-level candidates were first required to satisfy \(\mathrm{MHC}\geq0.90\) and \(\mathrm{TC}\geq0.90\), respectively, for every inference seed. When no candidate met this all-seed criterion, selection minimized the worst-seed coverage shortfall, followed by the mean shortfall and then the mean AFR. Such cases are marked by \(\dagger\). The exponentially weighted conformal baselines searched over \(\beta\), EnbPI over ensemble size, window size, and centering rule, ACI and AgACI over their update-rate grids, and CopulaCPTS over calibration split and training epochs.

For the CRC variants, all \(27\) prespecified multipliers were included in a single Holm correction using one-sided Hoeffding--Bentkus \(p\)-values, with no empirical fallback when formal certification was unavailable. Table~\ref{tab:crc-method-parameters} reports the calibrated quantities and selected LTT multipliers used in the final evaluation.

\begin{table}[!t]
\centering
\small
\setlength{\tabcolsep}{5pt}
\renewcommand{\arraystretch}{1.15}
\caption{Parameters used for the standard, adaptive, and structured conformal baselines.}
\label{tab:baseline-selected-parameters}
\begin{tabular}{p{3.1cm}p{11.2cm}}
\toprule
Method & Parameters used \\
\midrule

Global Residual
& Pooled split-conformal residual quantile, \(\alpha=0.10\). \\

Horizon-wise
& Per-horizon split-conformal residual quantiles, \(\alpha=0.10\). \\

Max-score Split
& Split conformal on
\(S_i=\max_{1\leq j\leq T}E_{i,j}\), \(\alpha=0.10\). \\

Residual Quantile
& \(50/50\) training/calibration split; QuantileRegressor with
quantile \(=0.90\), regularization \(=10^{-4}\), and solver = highs. \\

\midrule

EW-Horizon\(\,^\dagger\)
& \(\beta=0.02\); horizon-wise exponentially weighted quantiles with
online residual-history updates. \\

EW-Trajectory\(\,^\dagger\)
& \(\beta=0.02\); exponentially weighted trajectory-max quantiles with
online residual-history updates. \\

EnbPI-Horizon
& \(B=50\); full residual history; mean center; Ridge base model with
penalty \(=1\); online update. \\

EnbPI-Trajectory
& \(B=50\); window \(=100\); mean center; Ridge base model with
penalty \(=1\); online update. \\

ACI-Horizon\(\,^\dagger\)
& \(\gamma=0.05\); horizon-wise adaptive levels with online updates. \\

ACI-Trajectory\(\,^\dagger\)
& \(\gamma=0.02\); trajectory-max adaptive level with online updates. \\

AgACI-Horizon\(\,^\dagger\)
& \(\Gamma=\{0,0.005,0.01,0.02,0.03,0.05\}\);
\(\eta=0.05\); horizon-wise online aggregation. \\

AgACI-Trajectory\(\,^\dagger\)
& \(\Gamma=\{0,0.005,0.01,0.02,0.03,0.05\}\);
\(\eta=0.20\); trajectory-level online aggregation. \\

\midrule

Bonferroni-CRC
& Per-horizon correction
\(\alpha_j=\alpha/T=0.005\), with \(T=20\). \\

Sidak-CRC
& Per-horizon correction
\(\alpha_j=1-(1-\alpha)^{1/T}=0.005254\), with \(T=20\). \\

CopulaCPTS
& Quantile/copula calibration split \(=0.50/0.50\);
epochs \(=1200\); algorithmic random seed \(=123\). \\

\bottomrule
\end{tabular}
\end{table}

\begin{table}[H]
\centering
\small
\setlength{\tabcolsep}{6pt}
\renewcommand{\arraystretch}{1.15}
\caption{Validation performance of the selected baseline configurations, reported as mean \(\pm\) standard deviation across five diffusion inference seeds.}
\label{tab:baseline-validation-performance}
\begin{tabular}{lcccc}
\toprule
Method & Val. MHC & Val. TC & Val. fail & Val. AFR \\
\midrule

EW-Horizon\(\,^\dagger\)
& \(0.8977 \pm 0.0044\)
& \(0.7511 \pm 0.0151\)
& \(0.2489 \pm 0.0151\)
& \(10.9397 \pm 0.0901\) \\

EW-Trajectory\(\,^\dagger\)
& \(0.9760 \pm 0.0010\)
& \(0.8956 \pm 0.0054\)
& \(0.1044 \pm 0.0054\)
& \(19.9327 \pm 0.1515\) \\

EnbPI-Horizon
& \(0.9099 \pm 0.0071\)
& \(0.8200 \pm 0.0227\)
& \(0.1800 \pm 0.0227\)
& \(11.0966 \pm 0.0835\) \\

EnbPI-Trajectory
& \(0.9778 \pm 0.0009\)
& \(0.9044 \pm 0.0054\)
& \(0.0956 \pm 0.0054\)
& \(20.2111 \pm 0.1577\) \\

ACI-Horizon\(\,^\dagger\)
& \(0.8996 \pm 0.0014\)
& \(0.7600 \pm 0.0249\)
& \(0.2400 \pm 0.0249\)
& \(11.0983 \pm 0.2292\) \\

ACI-Trajectory\(\,^\dagger\)
& \(0.9771 \pm 0.0013\)
& \(0.8956 \pm 0.0054\)
& \(0.1044 \pm 0.0054\)
& \(20.1203 \pm 0.2402\) \\

AgACI-Horizon\(\,^\dagger\)
& \(0.8981 \pm 0.0042\)
& \(0.7756 \pm 0.0163\)
& \(0.2244 \pm 0.0163\)
& \(10.8788 \pm 0.1077\) \\

AgACI-Trajectory\(\,^\dagger\)
& \(0.9766 \pm 0.0015\)
& \(0.8956 \pm 0.0054\)
& \(0.1044 \pm 0.0054\)
& \(19.9937 \pm 0.2320\) \\

CopulaCPTS
& \(0.9780 \pm 0.0015\)
& \(0.9222 \pm 0.0141\)
& \(0.0778 \pm 0.0141\)
& \(14.4977 \pm 0.2848\) \\

\bottomrule
\end{tabular}
\end{table}

\begin{table}[H]
\centering
\small
\setlength{\tabcolsep}{4pt}
\renewcommand{\arraystretch}{1.15}
\caption{Method-specific parameters, calibrated quantities, and LTT outcomes for the conformal risk-control variants. Reported values are mean \(\pm\) standard deviation across five diffusion inference seeds; brackets show the observed range of \(\lambda^\star\).}
\label{tab:crc-method-parameters}
\resizebox{\textwidth}{!}{%
\begin{tabular}{p{3.2cm}p{4.8cm}p{5.0cm}p{3.2cm}}
\toprule
Method
& Fixed parameters
& Calibrated quantities
& LTT outcome \\
\midrule

Global-CRC
& No method-specific hyperparameters
& \(q_{\mathrm{global}}=11.607\pm0.170\)
& \(\lambda^\star=2.100\pm0.071\)
  \([2.000,2.200]\) \\

Horizon-Profile CRC
& Smoothing window \(K=3\);
  stabilization ratio \(\rho=0.15\)
& \(q_{\mathrm{global}}=11.607\pm0.170\)
& \(\lambda^\star=1.360\pm0.055\)
  \([1.300,1.400]\) \\

Trajectory-Stratified CRC
& Number of groups \(=2\);
  Ridge penalty \(\eta=1.0\)
& \(q_{0}^{\mathrm{raw}}=13.877\pm2.521\);

  \(q_{1}^{\mathrm{raw}}=23.850\pm1.907\)
& \(\lambda^\star=1.253\pm0.099\)
  \([1.167,1.400]\) \\

\raisebox{-2.5ex}[0pt][0pt]{\textbf{TRACE-CRC}}
& Number of groups \(=2\);
  smoothing window \(K=3\);
  stabilization ratio \(\rho=0.15\);
  Ridge penalty \(\eta=1.0\)
& \(q_{0}=10.857\pm0.386\);

  \(q_{1}=15.052\pm0.326\)
& \(\lambda^\star=1.073\pm0.037\)
  \([1.033,1.100]\) \\

\bottomrule
\end{tabular}%
}
\end{table}

\section{Robustness to the Internal Calibration Partition}
\label{app:robustness}

To assess the sensitivity of TRACE-CRC to the internal calibration partition, we fixed the outer \(300/700\) trajectory-level calibration/test split and repeated the evaluation over \(10\) random reallocations of the \(300\) calibration trajectories into
\(\mathcal D_{\mathrm{prof}}\), \(\mathcal D_{\mathrm{cp}}\), and
\(\mathcal D_{\mathrm{val}}\). Each partition was evaluated using five diffusion inference seeds, yielding \(50\) runs in total.
Table~\ref{tab:trace_crc_variability} summarizes the resulting variability. Coverage remained stable across runs, and the internal calibration partition had a larger effect than diffusion inference, although both sources of variability were modest.

\begin{table}[H]
\centering
\small
\caption{Variability of TRACE-CRC over \(50\) runs using \(10\) internal calibration partitions and five diffusion inference seeds. Internal-split and inference-seed SDs are marginal standard deviations across the corresponding averaged levels.}
\label{tab:trace_crc_variability}
\begin{tabular}{lccc}
\toprule
\textbf{Metric} & \textbf{Overall SD} & \textbf{Internal split SD} & \textbf{Inference-seed SD} \\
\midrule
MHC & 0.0062 & 0.0051 & 0.0021 \\
WHC & 0.0081 & 0.0063 & 0.0025 \\
TC  & 0.0155 & 0.0109 & 0.0058 \\
AFR & 0.9753 & 0.6691 & 0.3114 \\
\bottomrule
\end{tabular}
\end{table}
\end{document}